\documentclass{article} 
\usepackage{iclr2027_conference,times}

\usepackage{amsmath,amsfonts,bm}

\def\eqref#1{equation~\ref{#1}}

\def\1{\bm{1}}

\DeclareMathAlphabet{\mathsfit}{\encodingdefault}{\sfdefault}{m}{sl}
\SetMathAlphabet{\mathsfit}{bold}{\encodingdefault}{\sfdefault}{bx}{n}

\usepackage{graphicx}
\usepackage{hyperref}
\usepackage{url}
\usepackage{caption}
\usepackage{booktabs}
\usepackage{multirow, multicol, makecell, colortbl}

\title{Rethinking Generative Image Compression at Extremely Low Bitrates}

\author{%
\parbox[t]{\dimexpr\textwidth-2\tabcolsep\relax}{\raggedright
\mbox{\textbf{Tianyu Zhang}}\quad
\mbox{\textbf{Zhaoyang Jia}}\quad
\mbox{\textbf{Houqiang Li}}\quad
\mbox{\textbf{Dong Liu}} \\[5pt]
\textnormal{University of Science and Technology of China}
}%
}

\iclrfinalcopy
\begin{document}

\maketitle
\lhead{Preprint}

\begin{abstract}
Generative image compression produces visually plausible reconstructions at low bitrates, yet their behavior as the rate approaches zero remains largely unexplored. When pushed below normal operating rates, representative codecs undergo \textbf{semantic collapse}: rather than gracefully losing source-specific detail, they produce malformed or unrecognizable content. Our analysis identifies two factors. As the bitrate decreases, reconstruction losses increasingly conflict with semantic objectives on gradients and visual results, while pixel-space and reconstruction-oriented VAE diffusion models become less efficient on semantic preservation. Guided by these findings, we introduce \textbf{RAE-CoD}, a compression-oriented diffusion (CoD) built in a representation autoencoder (RAE) space with direct alignment between compressed and source representations, preserving recognizable, naturally structured content for a $256\times256$ image with as few as 16 bits. We evaluate this framework using five vision foundation models (VFM) and a blinded vision-language model protocol. On MSCOCO-30K, RAE-CoD stands out from all evaluation. At 0.001-0.008 bpp, it reduces relative VFM feature MSE and Fr\'echet Distance ratio by at least 25.7\% and 69.1\% over the best competitors. Meanwhile, semantic recognizability and quality of the reconstructions remain nearly constant while source consistency falls smoothly, replacing abrupt semantic collapse with a graceful transition toward unconditional generation. Code will be released at \url{https://github.com/LuizScarlet/RAE-CoD}.
\end{abstract}

\section{Introduction}
\label{sec:introduction}

Learned image compression ~\citep{balle2017end} steadily improves the efficiency of mapping images to entropy-coded latents. At low bitrates, however, minimizing distortion favors smooth averages that lack realistic details. Generative codecs address this limitation by incorporating adversarial training or diffusion priors to synthesize plausible reconstructions~\citep{mentzer2020high,careil2024perco}, yet existing evaluations nevertheless stop at a method-specific minimum rate. \textbf{What happens in the remaining interval between that operating point and zero bits is largely unknown.}

\begin{figure*}[t]
    \centering
    \includegraphics[width=\textwidth]{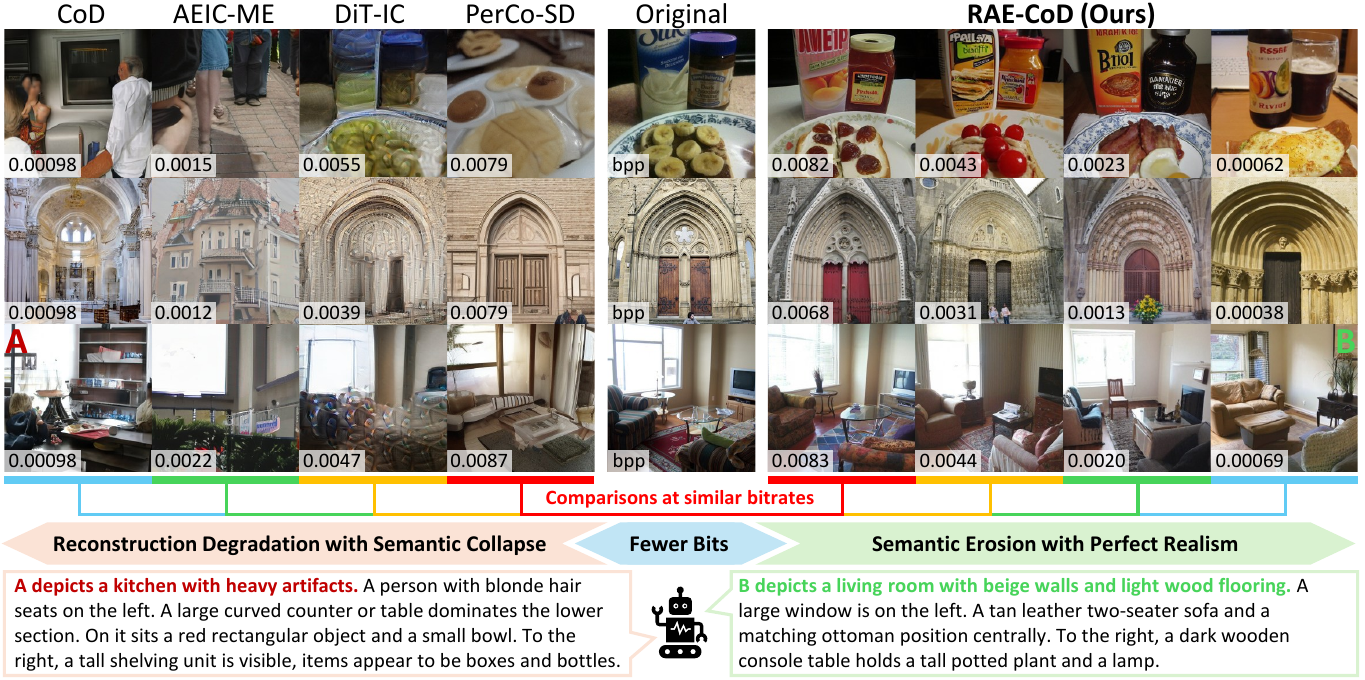}
    \caption{\textbf{Generative image compression below normal operating rates on MSCOCO-30K.} All methods are trained or finetuned and evaluated at $256\times256$. As the rate decreases, (\textbf{Left}) existing codecs exhibit reconstruction degradation followed by semantic collapse, whereas (\textbf{Right}) RAE-CoD maintains realistic and recognizable content while source consistency degrades more gradually. Captions are generated by Qwen3.5-9B. Numbers denote bits per pixel (bpp). \textit{Best viewed on screen.}}
    \label{fig:fig1}
\end{figure*}

This limiting regime exposes a distinction that conventional compression curves often obscure. As the rate approaches zero, a codec must lose source-specific information, but its output need not cease to be a coherent natural image, since an unconditional generative decoder can still sample recognizable objects and valid scenes even though they no longer correspond to the input. Ideally, semantic recognizability and naturalness should therefore remain high while source consistency degrades gradually. When we push representative generative codecs below their reported ranges, we instead observe an abrupt failure: objects deform, salient entities disappear, and scene structure becomes implausible, as illustrated in Fig.~\ref{fig:fig1}. We term this behavior \textbf{semantic collapse}. It suggests that existing codecs are not merely running out of bits. Their objectives and generative spaces fail to prioritize coherent semantic structure under an extreme bottleneck.

We investigate the factors separately. First, for reconstruction-driven codecs, which typically rely on reconstruction objectives like MSE and LPIPS~\citep{zhang2018lpips} to traverse the rate-distortion-perception tradeoff~\citep{blau2019rethinking}, we compare decoder feature gradients induced by MSE, LPIPS, and semantic similarities from vision foundation models. Reconstruction losses agree with one another but are nearly orthogonal to the semantic objectives, with negative alignment becoming more frequent at lower rates. Second, for diffusion-driven codecs, we use a zero-shot compression method~\citep{vonderfecht2025diffc} to compare pretrained diffusion transformers (DiT)~\citep{peebles2023scalable} established in pixel space, a reconstruction-oriented VAE space, and a representation autoencoder (RAE) space. Though pixel diffusion and VAE-based latent diffusion have been widely adopted in generative codecs~\citep{jia2026cod, li2024diffeic, zhang2025stablecodec}, both of them retains poor semantic preservation as the rate decreases. These findings motivate moving beyond reconstruction-centric supervision and diffusion modeling at extremely low bitrates.

Based on the analysis, we introduce \textbf{RAE-CoD}, a compression-oriented diffusion constructed in the semantic space of a RAE~\citep{zheng2025rae,singh2026raev2}. RAE-CoD exploits DINOv3~\citep{simeoni2025dinov3} to supply the clean representation target, while a latent codec fuses representation and pixels into entropy-coded latents. A conditional decoupled diffusion transformer (DDT)~\citep{wang2025ddt} then generates the source representation from the decoded condition. Instead of reconstruction objectives, we align the codec directly with the clean representation.

Evaluation at extremely low bitrates also requires separating whether an output is semantically well formed from whether it still depicts the source. We therefore introduce a two-level protocol. Vision foundation models (VFM) including CLIP, DINOv2, Inception-v3, SigLIP2, and ConvNeXt-v2~\citep{radford2021clip,oquab2023dinov2,szegedy2016inception,tschannen2025siglip2,woo2023convnextv2} measure paired feature similarity and distributional distance, while a metadata-blind vision-language model (VLM) judges semantic recognizability (SR), quality (SQ), and consistency (SC). On MSCOCO-30K ($256\times256$), RAE-CoD exhibits the strongest semantic fidelity in the evaluated range and maintains semantically coherent outputs down to 16 bits. Its SR and SQ remain nearly constant as the rate decreases, whereas SC falls smoothly, indicating the intended transition from source-conditioned generation toward unconditional generation. Our contributions include:
\begin{itemize}
\item We explore the behavior of generative image codecs between conventional operating ranges and zero bits. Specifically, we identify semantic collapse and analyze its causes through limitations of reconstruction-oriented training targets and diffusion spaces.
\item Guided by the analysis, we develop RAE-CoD, a compression-oriented diffusion constructed upon the representation space with direct alignment to the source representation, enabling semantic image coding with perfect realism towards 16 bits.
\item We combine multiple VFMs' evaluation with a blinded VLM protocol that disentangles semantic recognizability, quality, and consistency. Extensive comparisons show that RAE-CoD substantially improves semantics throughout the extremely low-rate scenarios.
\end{itemize}

\section{Generative Image Compression at Extremely Low Bitrates}
\label{sec:extreme_compression}

\begin{figure*}[t]
    \centering
    \includegraphics[width=0.9\textwidth]{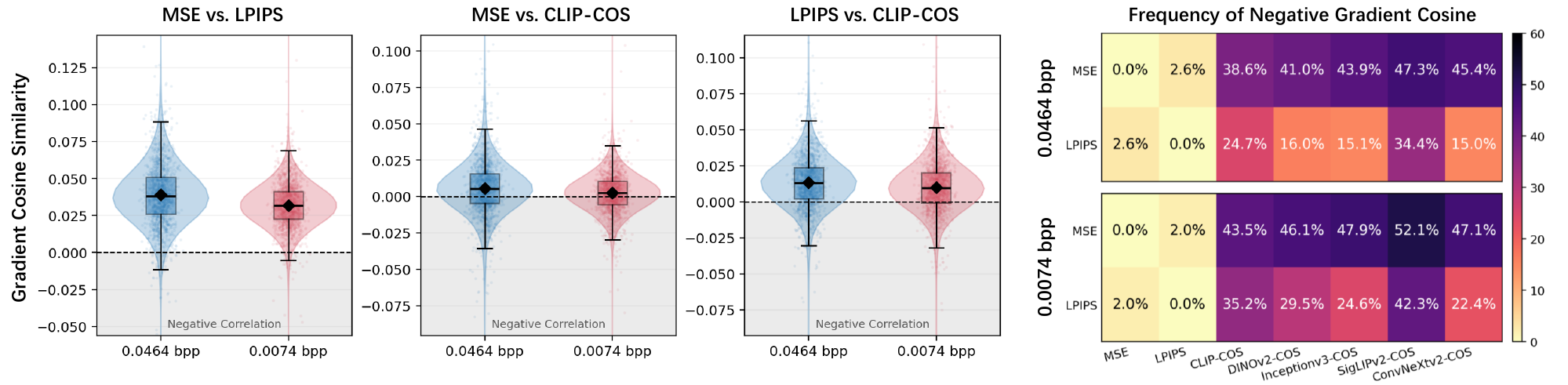}
    \caption{\textbf{Gradient geometry of reconstruction and semantic objectives.} We compute per-sample cosine similarities between loss gradients with respect to decoder features of a pretrained AEIC-ME on MSCOCO-30K. The first three panels show distributions for gradient pairs among MSE, LPIPS, and CLIP cosine similarity (CLIP-COS) at two rates. The right panel reports the frequency of negative gradient cosines between MSE/LPIPS and the cosine objectives of five VFMs.}
    \label{fig:objective_gradients}
\end{figure*}

\begin{figure*}[t]
    \centering
    \includegraphics[width=0.9\textwidth]{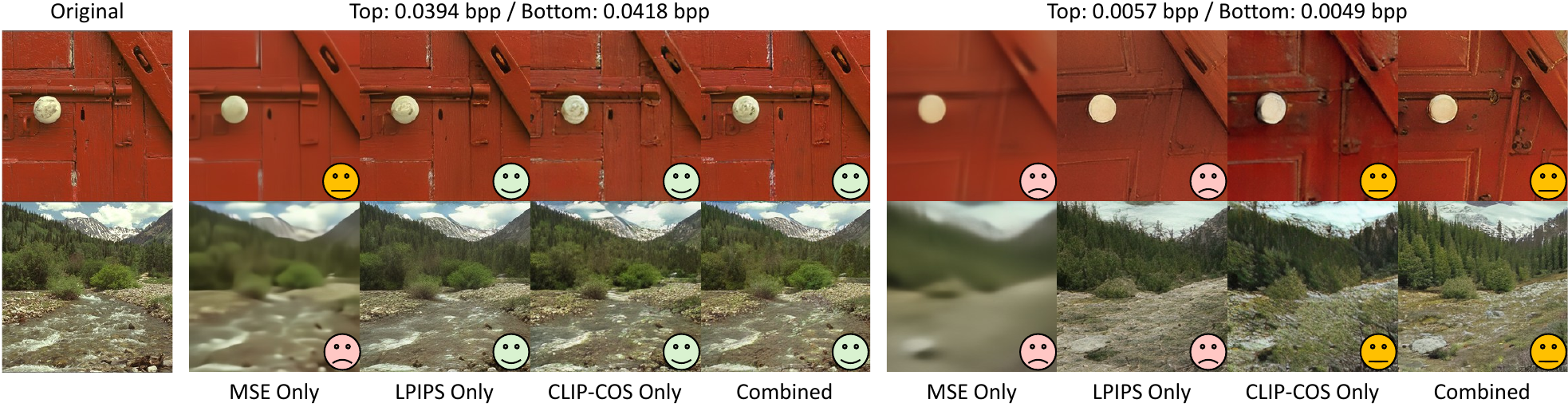}
    \caption{\textbf{Extreme test of decoder adaptation under different objectives.} We finetune pretrained AEIC-ME's decoder with MSE only, LPIPS only, CLIP-COS only, or their combination.}
    \label{fig:objective_visualization}
\end{figure*}

\subsection{What Happens at Extremely Low Bitrates?}
\label{sec:semantic_collapse}
Most generative codecs are evaluated only down to a method-specific minimum rate. We instead study the largely unexplored interval between this operating point and zero bits. According to the rate-distortion-perception tradeoff, source-specific information inevitably vanishes as the bitrate approaches zero. This does not, however, require the output itself to become semantically invalid. A zero-rate generative decoder can still sample from the natural-image distribution and attain perfect marginal realism, albeit without instance-level correspondence to the source. Empirically, pushing existing codecs below their reported operating ranges reveals a different failure. As shown in Fig.~\ref{fig:fig1}, recognizable objects deform into implausible structures, salient entities disappear, and scene meaning changes abruptly before the bitstream vanishes. We call this phenomenon \textbf{semantic collapse}, in which reconstructions retain neither source-consistent nor naturally formed semantic units.

\subsection{Analysis of Semantic Collapse}
\label{sec:semantic_collapse_analysis}
To investigate its cause, we group generative codecs into reconstruction- and diffusion-driven methods according to their dominant learning signal. We examine how reconstruction objectives and the intrinsic compression properties of diffusion spaces affect semantic preservation at extreme bitrates.

\begin{figure*}[t]
    \centering
    \includegraphics[width=0.9\textwidth]{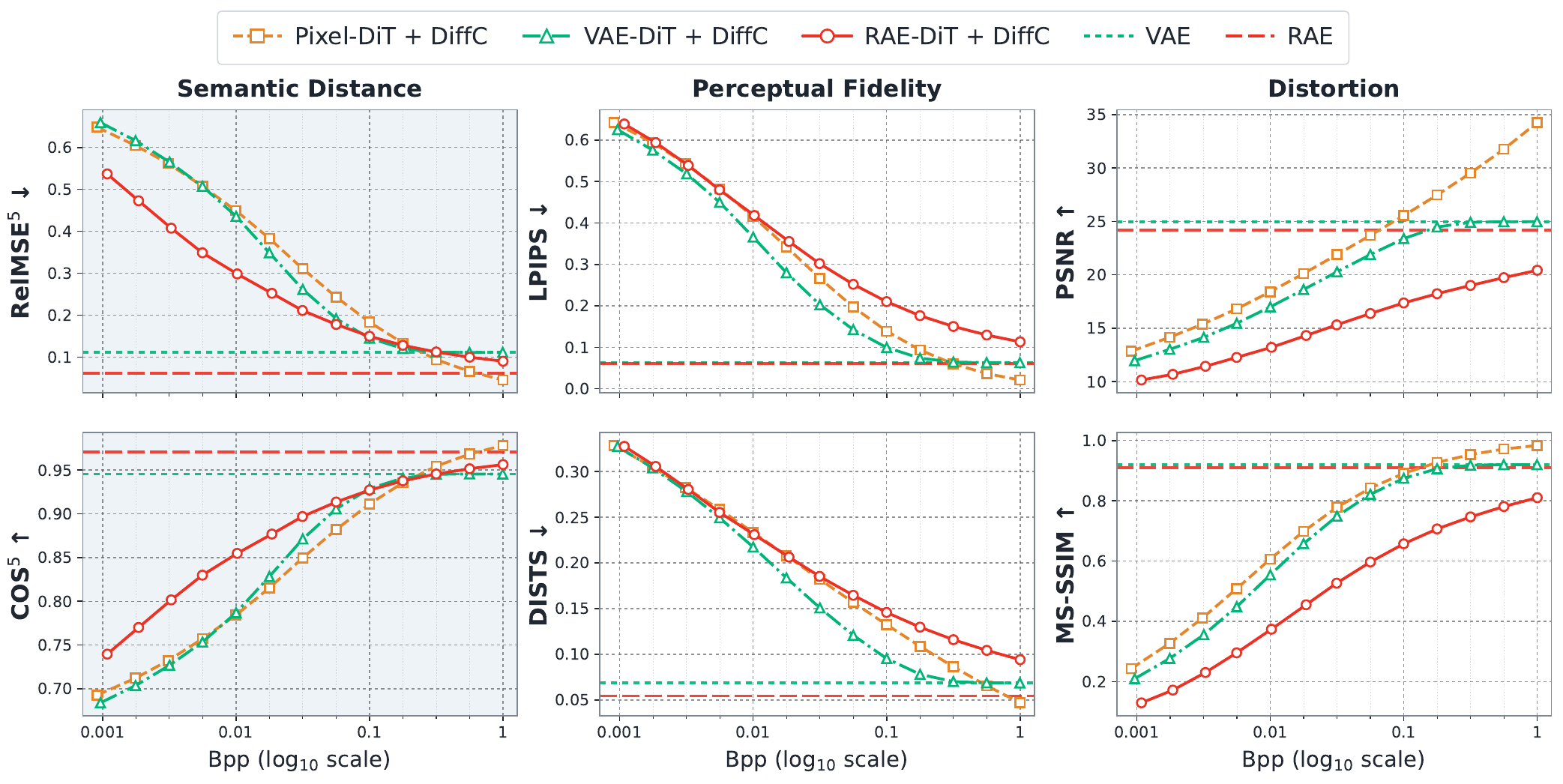}
    \caption{\textbf{Zero-shot compression with different diffusion models.} We apply the DiffC protocol to pretrained class-conditioned Pixel-DiT, VAE-DiT, and RAE-DiT on 5K ImageNet validation images at $256\times256$. All diffusion backbones use DDT-XL or comparable variants. Semantic distance is measured by the average relative MSE ($\mathrm{RelMSE}^{5}$) and cosine similarity ($\mathrm{COS}^{5}$) over five VFMs.}
    \label{fig:diffusion_space_quantitative}
\end{figure*}

\begin{figure*}[t]
    \centering
    \includegraphics[width=\textwidth]{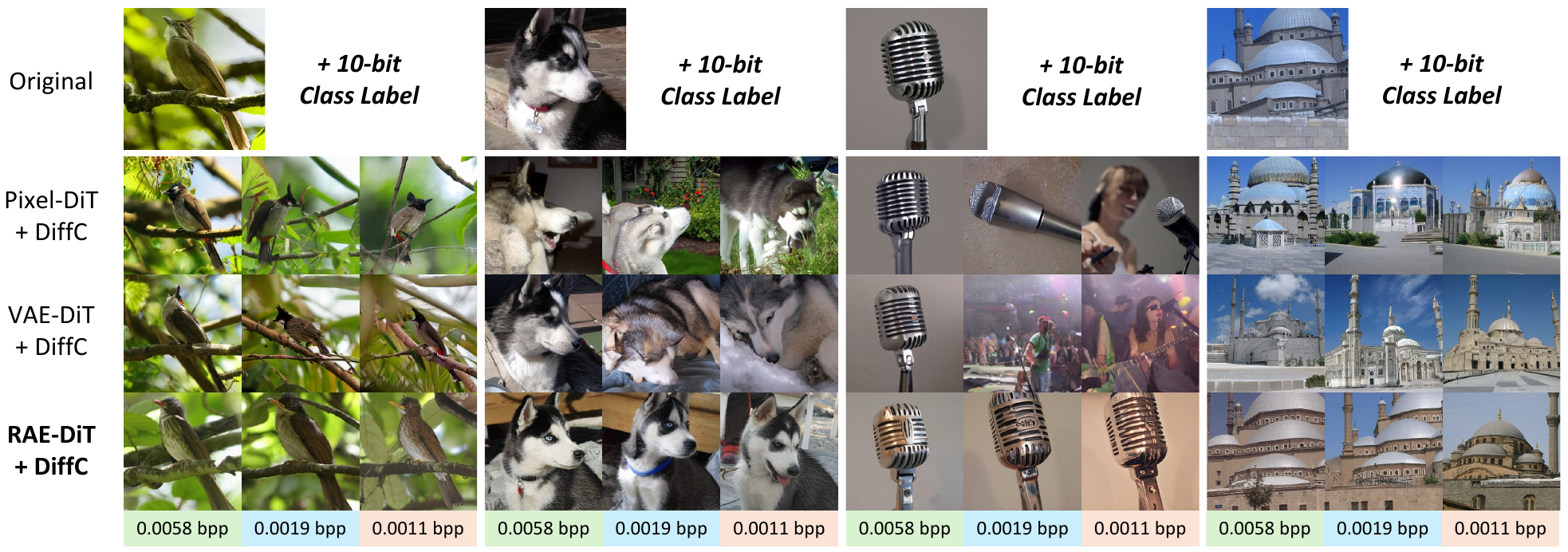}
    \caption{Qualitative DiffC comparison on ImageNet at $256\times256$.}
    \label{fig:diffusion_space_visualization}
\end{figure*}

\textbf{Directions of Reconstruction Objectives.} Reconstruction-driven methods, including GAN-based codecs and one-step diffusion codecs~\citep{mentzer2020high,zhang2025stablecodec,xue2025onedc}, exploit generative priors for perceptual compression but still rely heavily on reconstruction-oriented objectives, typically MSE and perceptual losses such as LPIPS. These losses help traverse the rate-distortion-perception trade-off, yet neither explicitly preserves the identities, relations, or global meaning of visible content. MSE penalizes pixel error, while LPIPS compares latents from a perceptual feature network. In Fig.~\ref{fig:objective_gradients}, we use the pretrained one-step diffusion codec AEIC-ME~\citep{zhang2026aeic} and analyze decoder feature gradients from seven losses: MSE, LPIPS, and cosine distances under CLIP, DINOv2, Inception-v3, SigLIP2, and ConvNeXt-v2. MSE and LPIPS rarely conflict with each other, but their gradients are nearly orthogonal to semantic objectives and become negatively aligned more frequently at the lower rate. The extreme test in Fig.~\ref{fig:objective_visualization} visualizes that MSE favors blur, while LPIPS produces structured artifacts as the bitrate decreases, both deviating entirely from semantic supervision which retains relatively better recognizable content. These findings indicate that reconstruction objectives do not reliably preserve semantics and can oppose semantic supervision, with this limitation becoming more consequential as the bit budget shrinks.

\textbf{Compression Properties of Diffusion Models.} Diffusion-driven methods~\citep{careil2024perco,jia2026cod,ke2025resulic} jointly optimize a codec and a conditional generative model in pixel or latent space. Their denoising loss learns a conditional posterior in the chosen data space, but does not make pixel coordinates or reconstruction-oriented VAE latents semantic by construction. We isolate the effect of that space using DiffC~\citep{vonderfecht2025diffc}, which performs reverse-channel coding with a pretrained diffusion model. Under the same protocol, we compare class-conditioned Pixel-DiT~\citep{ma2025deco}, VAE-DiT~\citep{wang2025ddt}, and RAE-DiT~\citep{singh2026raev2} models with comparable DDT-XL backbones. Fig.~\ref{fig:diffusion_space_quantitative} reveals a clear specialization: Pixel-DiT provides the strongest rate-distortion behavior and high-rate ceiling, VAE-DiT favors perceptual fidelity evaluated under LPIPS and DISTS~\citep{ding2020dists} at moderate rates. RAE-DiT, despite weaker distortion, preserves semantic representations most efficiently. Fig.~\ref{fig:diffusion_space_visualization} further shows that RAE-DiT retains coherent entities and scene layouts for substantially longer as the bitrate decreases. These results suggest that pixel-space and reconstruction-oriented VAE diffusion offer no intrinsic advantage for semantic preservation at extremely low rates.

\section{Compression-Oriented Diffusion with RAE}
\label{sec:method}

\begin{figure*}[t]
    \centering
    \includegraphics[width=0.95\textwidth]{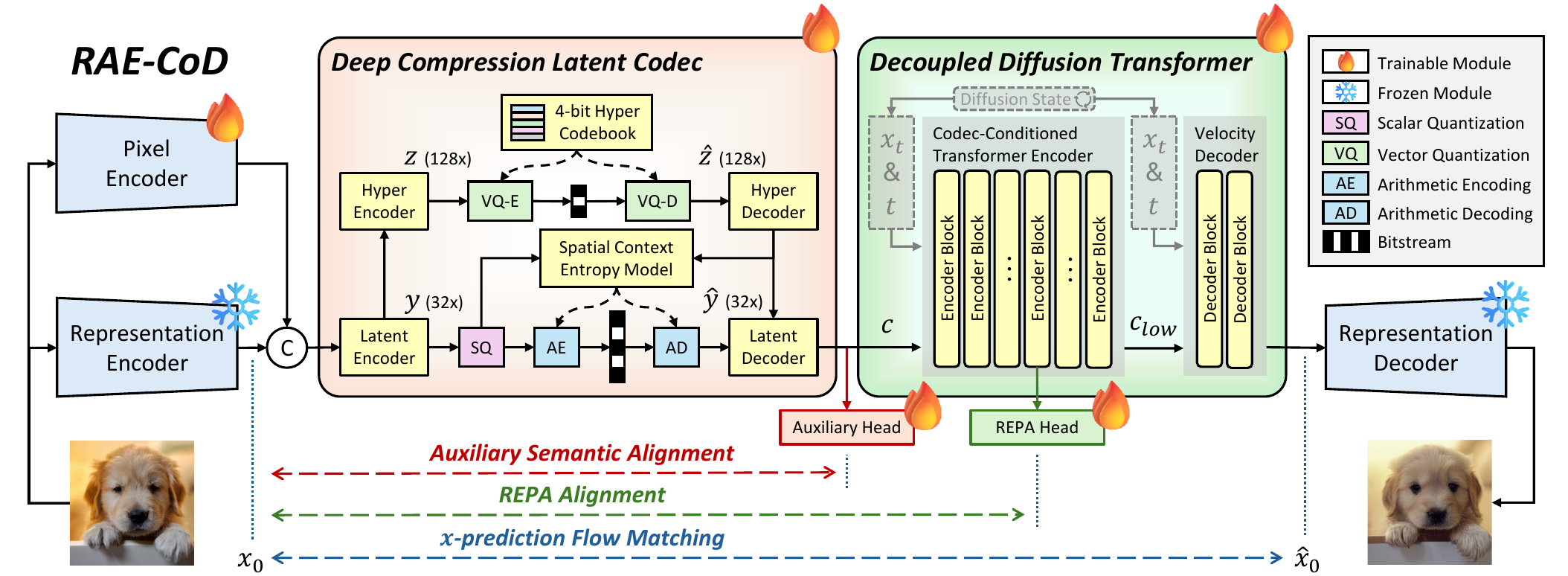}
    \caption{\textbf{Framework of RAE-CoD.} RAE-CoD comprises a frozen representation autoencoder, a deep compression latent codec, and a codec-conditioned decoupled diffusion transformer (DDT). All training objectives are performed in the semantically structured representation space.}
    \label{fig:method_pipeline}
\end{figure*}

The above analysis yields two findings for semantic collapse at extremely low bitrates. First, reconstruction objectives can conflict with semantics in their gradient directions. Second, diffusion in a representation space preserves semantics more efficiently as the bitrate decreases. In this section, we integrate both findings and introduce \textbf{RAE-CoD} (Fig.~\ref{fig:method_pipeline}), a compression-oriented diffusion model built with RAE and direct semantic condition alignment. It is designed to replace abrupt semantic collapse with a gradual loss of source consistency while retaining realistic, recognizable content.

\subsection{Pipeline}
\label{sec:method_pipeline}

\textbf{Representation Latent Space.} We adopt the DINOv3 encoder and its pretrained RAEv2 decoder~\citep{singh2026raev2}. Following the generalized RAE, we aggregate intermediate DINOv3 features, reshape the patch tokens spatially, and standardize them to obtain the clean diffusion target $x_0$ at $1/16$ of the input resolution. The representation decoder maps a generated $\hat{x}_0$ back to pixels.

\textbf{Deep Compression Latent Codec.} The latent codec places its entropy bottlenecks substantially deeper than standard neural codecs to support extremely low bitrates~\citep{zhang2025stablecodec}. A convolutional pixel encoder complements the frozen representation encoder with flexible pixel-level features. The codec concatenates their $1/16$-resolution outputs and produces a main latent $y$ at $1/32$ and a hyper latent $z$ at $1/128$. At extreme rates, the factorized model~\citep{balle2018variational} for $z$ can dominate the bits. We instead vector-quantize $z$ with a 4-bit (or 0.000244 bpp) codebook~\citep{esser2021vqgan}. The latent decoder reconstructs the $1/16$-resolution codec condition $c$ from $\hat{y}$ and $\hat{z}$.

\textbf{Conditional Decoupled Diffusion Transformer.} We initialize the denoiser from a pretrained RAEv2 DDT and replace its class-conditioning interface with the codec condition $c$. Following DDT~\citep{wang2025ddt}, the noisy representation tokens, timestep tokens, and projected codec tokens are concatenated and processed by a deep transformer encoder to produce a low-frequency self-condition $c_{low}$. A shallow but wide decoder head, reparameterized for $x$-prediction, uses $c_{low}$ to modulate the diffusion state and predict the clean representation $\hat{x}_0$.

\subsection{Diffusion Training Strategy}
\label{sec:diffusion_training}

\textbf{Loss Function.} We train RAE-CoD using $x$-prediction flow matching loss $\mathcal{L}_{\mathrm{FM}}$~\citep{liu2023rectifiedflow} and representation alignment. Following the RAE convention, a noisy latent is sampled along $x_t=(1-t)x_0+t\epsilon$, where $\epsilon\sim\mathcal{N}(0,I)$, and the full DDT predicts $x_0$ through $\mathcal{L}_{\mathrm{FM}}$. The early REPA head attached to the transformer encoder produces a second prediction $\hat{x}_0^{\mathrm{repa}}$. Because $x_0$ is itself the target vision representation, this head implements $\mathcal{L}_{\mathrm{REPA}}$~\citep{yu2025repa} as explicit early-layer $x$-prediction and also provides the weaker prediction used for internal guidance~\citep{zhou2026guiding}.

Although $\mathcal{L}_{\mathrm{FM}}$ and $\mathcal{L}_{\mathrm{REPA}}$ back-propagate through the codec condition, both are mediated by a randomly noised diffusion state and can provide an ambiguous learning signal when the codec is trained under a severe bottleneck. We therefore add a deterministic auxiliary objective $\mathcal{L}_{\mathrm{aux}}$ to the condition $c$. An auxiliary MLP head projects $c$, and $\mathcal{L}_{\mathrm{aux}}$ measures the cosine distance relative to the clean target $x_0$, requiring the entropy-constrained $c$ to preserve semantics~\citep{Zhang_2026_CVPR}. Given the entropy $\mathcal{R}_{y}$ and the commitment loss $\mathcal{L}_{\mathrm{VQ}}$~\citep{esser2021vqgan}, the complete objective is
\begin{equation}
    \mathcal{L}=
    \lambda_{\mathrm{rate}}\mathcal{R}_{y}
    +\lambda_{\mathrm{vq}}\mathcal{L}_{\mathrm{VQ}}
    +\mathcal{L}_{\mathrm{FM}}
    +\lambda_{\mathrm{repa}}\mathcal{L}_{\mathrm{REPA}}
    +\lambda_{\mathrm{aux}}\mathcal{L}_{\mathrm{aux}}.
    \label{eq:overall_loss}
\end{equation}

\textbf{Progressive Training towards Extremely Low Bitrates.} Directly imposing a severe rate penalty can destroy the pretrained diffusion prior. Inspired by implicit bitrate pruning~\citep{zhang2025stablecodec}, we progressively increase $\lambda_{\mathrm{rate}}$ in training Stage~I while applying LoRA~\citep{hu2022lora} for the pretrained DDT. In Stage~II, we branch from the corresponding Stage-I checkpoints, fix $\lambda_{\mathrm{rate}}$ at each target value, merge the LoRA weights, and fine-tune all DDT parameters jointly with the codec.

\textbf{Implementation.} We train on 23.2M $256\times256$ images from ImageNet-21K~\citep{russakovsky2015imagenet} and CC12M~\citep{changpinyo2021cc12m}. We set $\lambda_{\mathrm{repa}}=1$, $\lambda_{\mathrm{vq}}=0.25$, and $\lambda_{\mathrm{aux}}=0.5$. In Stage~I, we use rank-32 LoRA and increase $\lambda_{\mathrm{rate}}$ from 0.1 through $\{2,12,16,24,32,48\}$ at 30K-iteration intervals with a learning rate of $10^{-4}$. In Stage~II, we fix $\lambda_{\mathrm{rate}}\in\{12,16,24,32,48\}$ and train each operating point for 100K iterations at $10^{-5}$. Both stages use AdamW, an effective batch size of 128, bfloat16 mixed precision, and an EMA with decay 0.9995 on four NVIDIA A100 GPUs. 

\section{Evaluation Protocol}
\label{sec:semantic_evaluation}
Pixel distortion and perceptual fidelity, though commonly adopted in evaluating generative image compression, fail to distinguish the semantic differences at extremely low bitrates identified in Fig.~\ref{fig:objective_visualization} and \ref{fig:diffusion_space_quantitative}. We therefore introduce a two-level evaluation protocol that combines deterministic vision foundation model (VFM) features with a blinded vision-language model (VLM) judge.

\textbf{VFM-Based Evaluation.} Let $\Phi=\{\phi_m\}_{m=1}^{5}$ contain Inception-v3, ConvNeXt-v2, DINOv2, SigLIP2, and CLIP, spanning supervised, self-supervised, and vision-language objectives. For each source-reconstruction pair, we compute feature MSE and cosine similarity to measure semantic distance, and report the average relative MSE ($\mathrm{RelMSE}^{5}$) and cosine similarity ($\mathrm{COS}^{5}$) for aggregation across five VFMs. For distributional distance, we compute Fr\'echet Distance (FD) on different VFMs, and adapts the FD ratio~\citep{yang2026fdloss} to compression. Let $\mathcal{X}$ denote the source images, $\hat{\mathcal{X}}$ the reconstructions for evaluation, and $\hat{\mathcal{X}}_{\mathrm{A}}$ the reconstructions of an anchor codec. We define
\begin{equation}
    \mathrm{FDr}^{5}(\hat{\mathcal{X}})=\frac{1}{5}\sum_{m=1}^{5}
    \frac{\mathrm{FD}_{\phi_m}(\mathcal{X},\hat{\mathcal{X}})}
         {\mathrm{FD}_{\phi_m}(\mathcal{X},\hat{\mathcal{X}}_{\mathrm{A}})}.
    \label{eq:fdr5}
\end{equation}

\textbf{VLM-Based Evaluation.} Features can overlook failures that are obvious to a human, such as malformed objects and changed relations. We therefore complement them with a blinded Qwen3.5-9B judge~\citep{qwen2026qwen35} and evaluate three distinct properties. Semantic recognizability (SR) asks what content can be identified from the reconstruction alone. Semantic quality (SQ) asks whether that recognizable content is coherent, structurally intact, natural, and not dominated by artifacts. Semantic consistency (SC) instead asks how faithfully the meaning of the source is retained.

To compute SR and SQ, the VLM sees only an anonymous reconstruction and decomposes its visible content into nonredundant semantic units, such as entities, actions, and relations. For each unit $c$, it assigns semantic importance $w_c$, identity confidence $p_c$, and intrinsic quality $q_c$. The raw scores $\mathrm{SR}^{\mathrm{raw}}$ and $\mathrm{SQ}^{\mathrm{raw}}$ are importance-weighted averages. Since VLM does not use the raw $[0,100]$ scale uniformly across images, we normalize $\mathrm{SR}^{\mathrm{raw}}$ and $\mathrm{SQ}^{\mathrm{raw}}$ using per-image controls. For metric $\mathrm{M}\in\{\mathrm{SR},\mathrm{SQ}\}$, the clean source provides an upper anchor $\mathrm{U}^\mathrm{M}$, while a minimum-quality JPEG compression of that source provides a lower anchor $\mathrm{L}^\mathrm{M}$. The final scores are computed as:
\begin{equation}
    \mathrm{SR}^{\mathrm{raw}}=
    \frac{\sum_{c}w_cp_c}{\sum_{c}w_c},
    \quad
    \mathrm{SQ}^{\mathrm{raw}}=
    \frac{\sum_{c}w_cq_c}{\sum_{c}w_c},
    \quad
    \mathrm{M}=100 \cdot \operatorname{\textbf{clip}}\left[
    \frac{\mathrm{M}^{\mathrm{raw}}-\mathrm{L}^\mathrm{M}}{\mathrm{U}^\mathrm{M}-\mathrm{L}^\mathrm{M}},0,1\right].
\end{equation}

To compute SC, the VLM independently inventories the source; the two inventories are then matched in a separate text-only call, without using the SR/SQ judgments. This produces semantic recall $\rho$, which measures retained source content, and semantic precision $\pi$, which penalizes invented or substituted content. Their harmonic mean $F$ is combined with global scene similarity $G$:
\begin{equation}
    F=2\rho\pi / (\rho+\pi),
    \qquad
    \mathrm{SC}=0.4 \cdot G+0.6 \cdot F.
\end{equation}
\section{Experiments}
\label{sec:experiments}

\begin{figure*}[t]
    \centering
    \includegraphics[width=\textwidth]{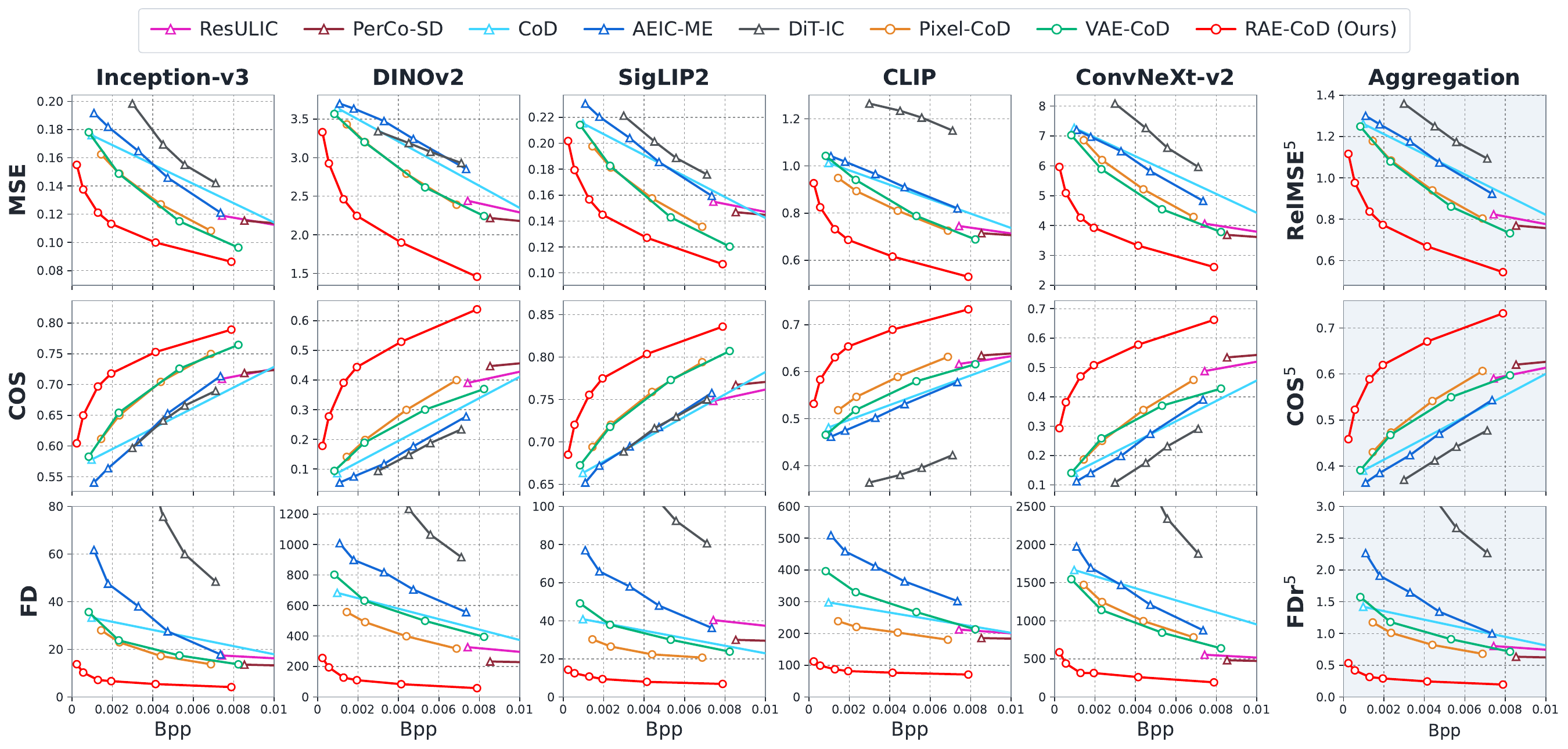}
    \caption{VFM-based semantic and quality evaluation on MSCOCO-30K.}
    \label{fig:rd}
\end{figure*}

\begin{figure*}[t]
    \centering
    \includegraphics[width=\textwidth]{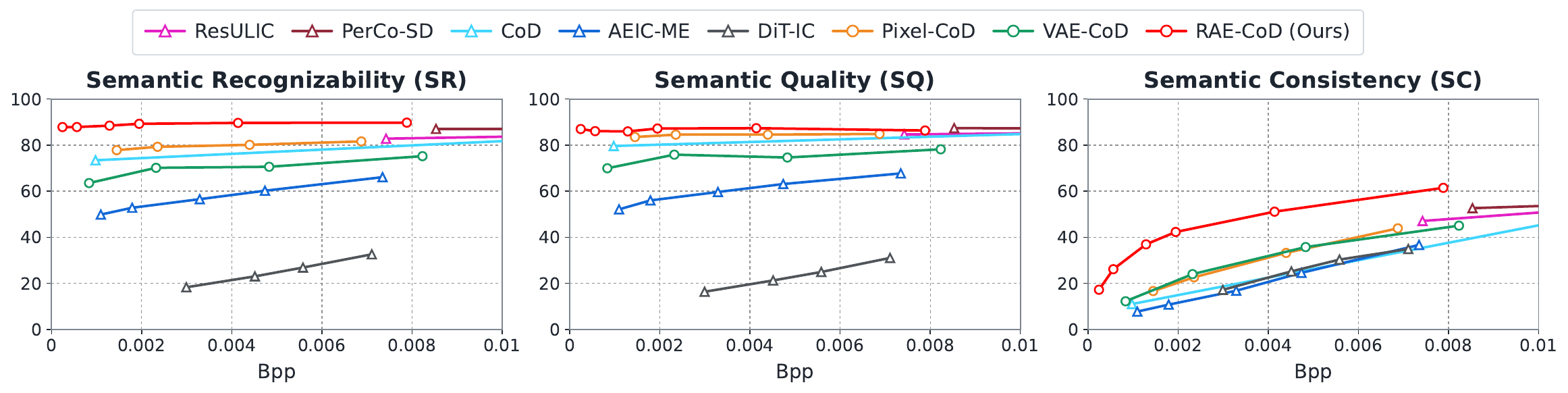}
    \caption{VLM-based semantic and quality evaluation on MSCOCO-30K.}
    \label{fig:rd_vlm}
\end{figure*}

\begin{figure*}[t]
    \centering
    \includegraphics[width=\textwidth]{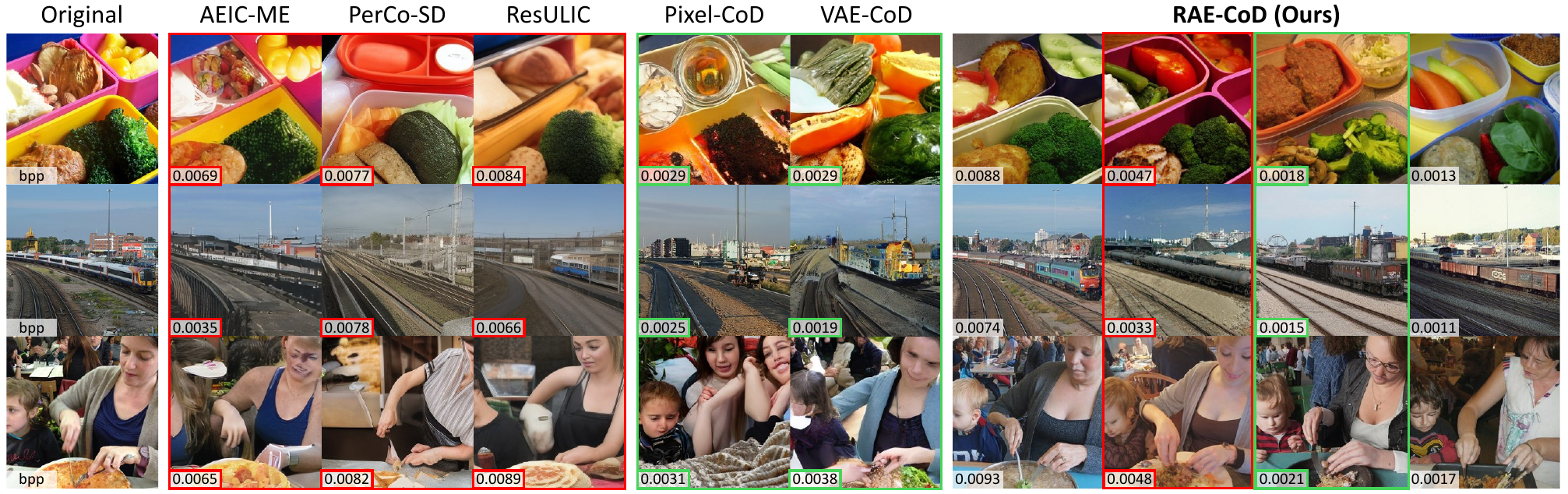}
    \caption{\textbf{Qualitative comparison on MSCOCO-30K at $256\times256$.} Red and green borders highlight comparisons at similar bitrates. RAE-CoD is shown in increasingly aggressive bitrates.}
    \label{fig:visual}
\end{figure*}

\begin{figure*}[t]
\centering
\begin{minipage}[t]{0.4\textwidth}
    \centering
    \includegraphics[width=\linewidth]{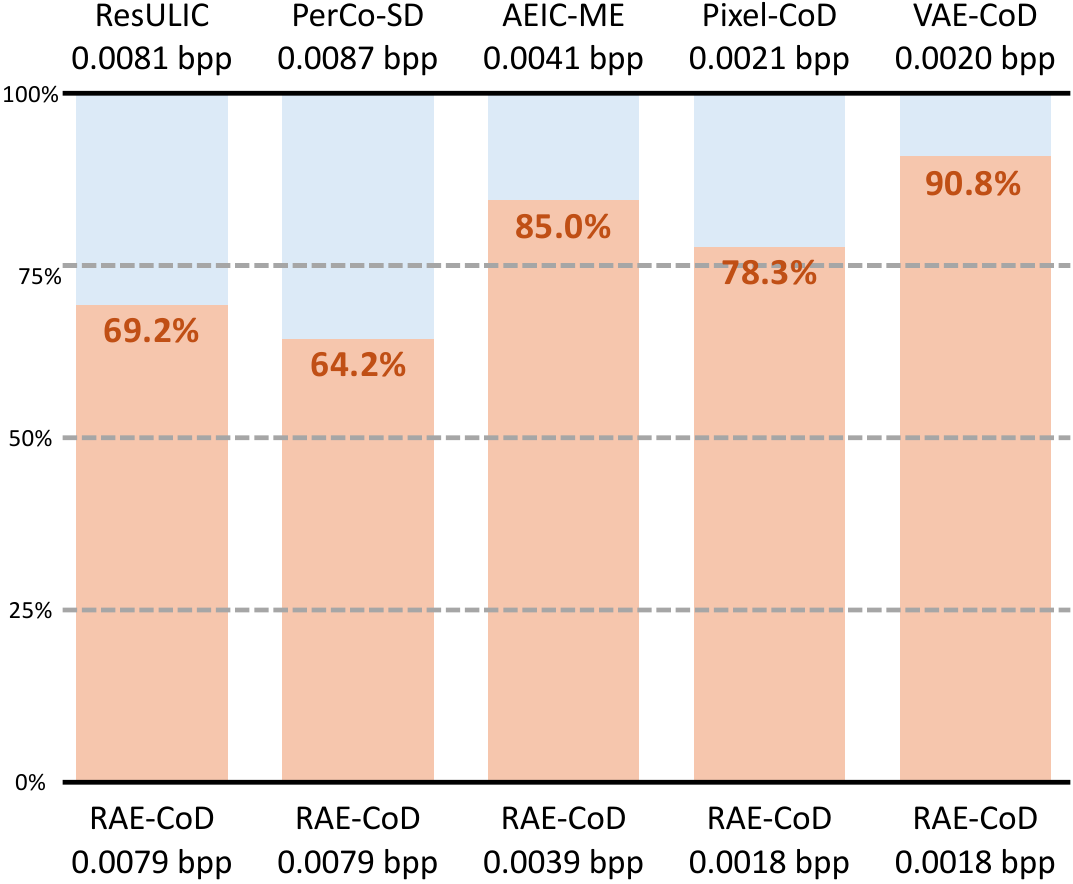}
    \caption{User study on Kodak.}
    \label{fig:human}
\end{minipage}\hfill
\begin{minipage}[t]{0.57\textwidth}
    \centering
    \includegraphics[width=\linewidth]{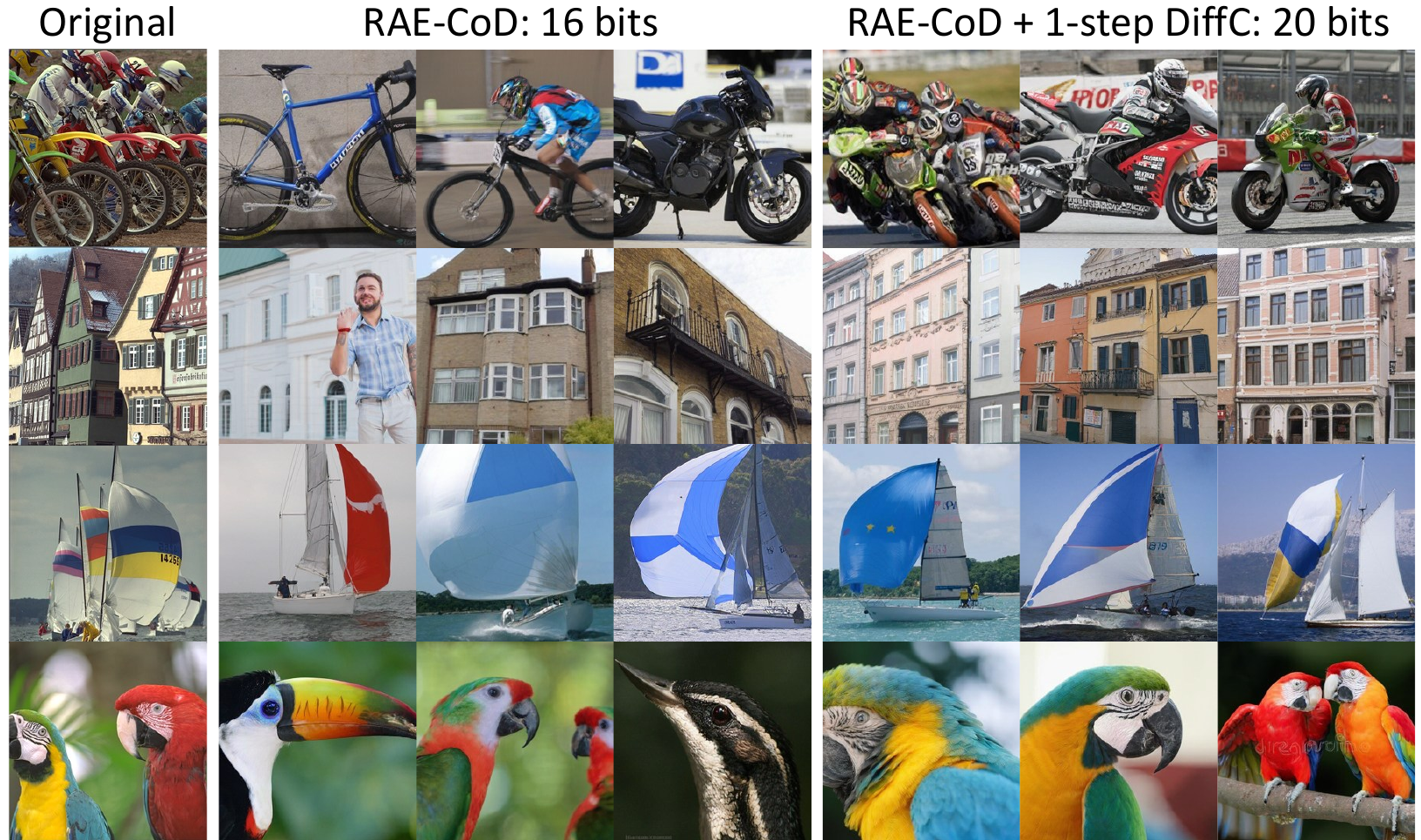}
    \caption{16-bit RAE-CoD on Kodak at $256\times256$.}
    \label{fig:16bit}
\end{minipage}
\end{figure*}

\subsection{Settings}
\label{sec:experimental_settings}

\textbf{Compared Methods.} We compare with five representative generative codecs: PerCo-SD~\citep{korber2024percosd}, ResULIC~\citep{ke2025resulic}, CoD~\citep{jia2026cod}, AEIC-ME~\citep{zhang2026aeic}, and DiT-IC~\citep{shi2026ditic}. To reach extreme rates, we start from the lowest-rate official checkpoints and finetune them at $256\times256$ using their code. We also construct two space-controlled variants. Pixel-CoD replaces RAE-DiT with a pretrained pixel-space DDT~\citep{ma2025deco}, while VAE-CoD uses a pretrained VAE-space DDT~\citep{wang2025ddt}; both retain our deep latent codec and training strategy. Pixel-CoD differs from CoD~\citep{jia2026cod} mainly in using a pretrained DeCo rather than training from scratch, and using an entropy model instead of a VQ bottleneck.

\textbf{Test Data and Evaluation.} We primarily use MSCOCO-30K~\citep{careil2024perco,lin2014coco} for all evaluation, a 30K-image subset of commonly used in image compression. We also adopt Kodak~\citep{kodak1999}. All images are resized and center-cropped to $256\times256$. We report theoretical bits per pixel (bpp) estimated from the learned likelihoods. Following Sec.~\ref{sec:semantic_evaluation}, the VFM evaluation measures features MSE, COS, and FD under five VFMs. We aggregate them as $\mathrm{RelMSE}^{5}$, $\mathrm{COS}^{5}$, and $\mathrm{FDr}^{5}$. $\mathrm{FDr}^{5}$ is normalized by the corresponding highest-rate AEIC-ME value. We further apply the Qwen3.5-9B protocol to report SR, SQ, and SC.

\subsection{Main Results}
\label{sec:main_results}

\textbf{Quantitative Comparisons.} Fig.~\ref{fig:rd} shows that RAE-CoD establishes the strongest semantic envelope throughout the evaluated range, on both aggregated results and specific VFMs. Around 0.008 bpp, RAE-CoD obtains a 25.7\% reduction in $\mathrm{RelMSE}^{5}$ and a 69.1\% reduction in $\mathrm{FDr}^{5}$ relative to the best competing value, while the advantage grows in the more challenging region near 0.001 bpp. The consistent gap over Pixel-CoD and VAE-CoD isolates the benefit of constructing compression-oriented diffusion in the RAE space. The VLM results in Fig.~\ref{fig:rd_vlm} reveal the behavior hidden by a single similarity score. Across the full RAE-CoD curve, SR remains within 86.7--87.2 and SQ within 69.2--70.7, while SC decreases from 61.5 to 17.2. This is the desired behavior: reconstructions remain recognizable and semantically well formed, but progressively less tied to the source.

\textbf{Human Preference Study and Qualitative Comparisons.} Fig.~\ref{fig:human} reports pairwise preferences on Kodak, with 120 votes from 10 participants for each comparison. At similar bitrates, RAE-CoD is preferred in all five comparisons against ResULIC, PerCo-SD, AEIC-ME and CoD variants, receiving 64.2-90.8\% of the votes. The examples in Fig.~\ref{fig:visual} show the consistent trend. RAE-CoD better preserves the salient content and remains visually coherent as the rate decreases, while competing methods lose source entities or develop malformed structures.

\subsection{Discussion and Ablation Study}
\label{sec:discussion_ablation}

\textbf{16-Bit Compression.} At the lowest operating point, we drop the main latent and finetuning the model sorely relying on the 16-bit hyper latent. Fig.~\ref{fig:16bit} shows that this 16-bit message is already sufficient to bias the generation toward the broad scene semantics, but different noise samples can largely change object identity or layout details at this bitrate. We therefore test a lightweight improvement using a single DiffC reverse-channel-coding step~\citep{vonderfecht2025diffc}. Communicating around four additional bits (as theoretically estimated) constrains the sampled diffusion state and improves semantic stability, while the total rate remains only 20 bits, or 0.000305 bpp.

\begin{table*}[t]
\centering
\setlength{\tabcolsep}{2pt}
\renewcommand{\arraystretch}{1.1}
\begin{minipage}[t]{0.48\textwidth}
\caption{\textbf{Ablation on the auxiliary losses.} BD-rate is computed over four points from 0.001 to 0.008 bpp relative to no auxiliary loss.}
\centering
\footnotesize
\begin{tabular}{l|ccc}
\toprule
\multirow{2}{*}{\makecell{Auxiliary Objective}} & \multicolumn{3}{c}{$\,\,\,\,\,\,\,$BD-rate ($\downarrow\%$)} \\
 & $\,\mathrm{RelMSE}^{5}$ & $\,\mathrm{COS}^{5}$ & $\,\mathrm{FDr}^{5}$ \\
\midrule
None & 0 & 0 & 0 \\
Pixel MSE only & $-43.38$ & $-41.75$ & $-73.20$ \\
\rowcolor{gray!20}
Semantic only & $\mathbf{-76.87}$ & $\mathbf{-77.18}$ & $\mathbf{-87.49}$ \\
Both & $-65.58$ & $-64.90$ & $-81.98$ \\
\bottomrule
\end{tabular}

\label{tab:ablation_loss}
\end{minipage}\hfill
\renewcommand{\arraystretch}{1.32}
\begin{minipage}[t]{0.48\textwidth}
\caption{\textbf{Ablation on the encoders.} BD-rate is computed over four points from 0.001 to 0.008 bpp relative to RAE encoder only.}
\centering
\footnotesize
\begin{tabular}{l|ccc}
\toprule
\multirow{2}{*}{\makecell{Encoder Selection}} & \multicolumn{3}{c}{$\,\,\,\,\,\,\,$BD-rate ($\downarrow\%$)} \\
 & $\,\mathrm{RelMSE}^{5}$ & $\,\mathrm{COS}^{5}$ & $\,\mathrm{FDr}^{5}$ \\
\midrule
RAE encoder only & 0 & 0 & 0 \\
Pixel encoder only & $+50.65$ & $+49.47$ & $+37.35$ \\
\rowcolor{gray!20}
Both & $\mathbf{-32.10}$ & $\mathbf{-32.92}$ & $\mathbf{-66.34}$ \\
\bottomrule
\end{tabular}

\label{tab:ablation_encoder}
\end{minipage}
\end{table*}

\textbf{Auxiliary Codec Alignment.} Table~\ref{tab:ablation_loss} compares the losses applied to the codec condition using BD-rate~\citep{bjontegaard2001bdrate}, where we test pixel-level MSE as the reconstruction loss. Semantic alignment alone yields substantially stronger performance, which agrees with Fig.~\ref{fig:objective_gradients} and ~\ref{fig:objective_visualization} that reconstruction supervision can interferes with the semantic signal under an extreme bitrate bottleneck.

\textbf{Selection of Encoders.} Table~\ref{tab:ablation_encoder} studies the inputs for the latent codec. Using only the pixel encoder requires at least 37.35\% more bits than using only the RAE encoder at matched aggregate quality, confirming that representation features provide the more compression-efficient semantic basis. Nevertheless, the two encoders are complementary as fusing pixel and RAE features reduces bitrate by 32.10\% under $\mathrm{RelMSE}^{5}$, 32.92\% under $\mathrm{COS}^{5}$, and 66.34\% under $\mathrm{FDr}^{5}$ relative to RAE-E alone.

\section{Related Work}

\textbf{Diffusion-based Generative Image Compression.} Diffusion-based codecs introduce learned priors to improve realism, broadly involving reverse-channel coding with a pretrained diffusion~\citep{theis2022lossy,vonderfecht2025diffc}, generative refinement~\citep{ghouse2023dirac,hoogeboom2023high}, and end-to-end training of a conditional diffusion decoder~\citep{yang2023lossy}. Ultra-low-rate variants further communicate side information and compressed  controls~\citep{lei2023textsketch,careil2024perco,li2024diffeic,ke2025resulic}, while recent work emphasizes one-step generation, lightweight coding, and compression-specific pretraining~\citep{,guo2025oscar,zhang2025stablecodec,xue2025onedc,jia2026cod,zhang2026aeic,jia2026codlite}. In contrast, we explore the behavior of representative generative codecs when pushed below their operating bitrates.

\textbf{Vision Representation in Image Generation.} Self-supervised and language-supervised vision encoders provide richer semantic and spatial structure~\citep{rombach2022ldm,he2022mae,oquab2023dinov2,radford2021clip}. Image generation exploits these representations by aligned denoiser features~\citep{yu2025repa,leng2025repae,singh2025irepa} and redesigned generative latent through semantic regularization, masked representation learning or frozen representation encoders~\citep{yao2025vavae,xu2025reals,chen2025maetok,zheng2025rae,gao2025fae,singh2026raev2}. These trends establish representation structure as central to generation, we instead study their ability to preserve semantics for generative image compression at extreme bitrates.

\section{Conclusion}
\label{sec:conclusion}

We studied generative image compression in the largely unexplored interval between conventional operating rates and zero bits. Representative codecs do not always lose source information gracefully in this regime. Instead, they undergo semantic collapse. Our analysis connected this failure to the weak alignment between reconstruction and semantic objectives and to the limited semantic efficiency of pixel and reconstruction-oriented VAE diffusion spaces. These observations led to RAE-CoD, which performs compression-oriented diffusion in a pretrained representation space and directly aligns its compressed condition with the source representation. Together with our VFM and VLM evaluation protocol, the experiments show that RAE-CoD preserves recognizable, naturally structured content down to 16 bits while allowing source consistency to decrease gradually. We hope these findings encourage future work to push the lower bitrate frontier of generative compression.

\textbf{Limitations.} Our experiments are primarily limited to $256\times256$ images and a 0.9B-parameter decoupled diffusion transformer due to computational constraints. We therefore do not establish a scaling law: it remains unclear how larger generative models built in the representation space can further improve the semantic stability and quality at extremely low bitrates, or whether high-resolution image semantics can be preserved with similar or even smaller bitstreams. Exploring model and resolution scaling is an important direction for future work.

\clearpage

\bibliography{reference}
\bibliographystyle{iclr2027_conference}

\clearpage
\appendix

\section{Discussion on Vision Foundation Models}
\label{sec:app_vfm}

\subsection{Model Configurations and Evaluation}
\label{sec:app_vfm_config}

The five vision foundation models (VFMs) used in Sec.~\ref{sec:semantic_evaluation} were selected to cover different architectures and pretraining signals rather than to rely on one notion of visual similarity. Table~\ref{tab:app_vfm_config} gives the exact configurations. All networks are frozen. Images are loaded in RGB, converted to the value range of $[0,1]$, resized with the preprocessing associated with each checkpoint, and normalized using its pretrained statistics. For the transformer encoders, we extract one global token from the final layer; for the convolutional encoders, we average the final spatial feature map.

\begin{table*}[t]
\centering
\caption{VFM configurations used for evaluation.}
\label{tab:app_vfm_config}
\footnotesize
\setlength{\tabcolsep}{2.5pt}
\begin{tabular}{l l l c c}
\toprule
Model & Checkpoint identifier & Arch. & Input & Dim. \\
\midrule
Inception-v3~\citep{szegedy2016inception}
& \texttt{inception\_v3} & CNN & $299^2$ & 2048 \\
ConvNeXt-v2~\citep{woo2023convnextv2}
& \texttt{convnextv2\_base.fcmae\_ft\_in22k\_in1k} & CNN & $224^2$ & 1024 \\
DINOv2~\citep{oquab2023dinov2}
& \texttt{vit\_large\_patch14\_dinov2.lvd142m} & ViT & $256^2$ & 1024 \\
SigLIP2~\citep{tschannen2025siglip2}
& \texttt{vit\_so400m\_patch16\_siglip\_256.v2\_webli} & ViT & $224^2$ & 1152 \\
CLIP~\citep{radford2021clip}
& \texttt{vit\_large\_patch14\_clip\_224.openai} & ViT & $256^2$ & 1024 \\
\bottomrule
\end{tabular}
\end{table*}

For source $x_i$, reconstruction $\hat{x}_i$, and encoder $\phi_m$, the implementation computes
\begin{equation}
 \mathrm{RelMSE}_{m,i}=\frac{\operatorname{mean}[(\phi_m(x_i)-\phi_m(\hat{x}_i))^2]}
 {\operatorname{mean}[\phi_m(x_i)^2]+\epsilon},
 \qquad
 \mathrm{COS}_{m,i}=\frac{\phi_m(x_i)^\top\phi_m(\hat{x}_i)}
 {\lVert\phi_m(x_i)\rVert_2\lVert\phi_m(\hat{x}_i)\rVert_2}.
 \label{eq:app_vfm_metrics}
\end{equation}
We average $\mathrm{RelMSE}_{m,i}$ and $\mathrm{COS}_{m,i}$ first over images and then equally over the five models to obtain $\mathrm{RelMSE}^{5}$ and $\mathrm{COS}^{5}$. Fr\'echet Distance~\citep{yang2026fdloss} (FD) is computed independently in each feature space from means and covariances. As described in Sec.~\ref{sec:semantic_evaluation}, the five distances are normalized separately before being averaged into $\mathrm{FDr}^{5}$. Per-model normalization and equal averaging prevent a representation with a larger dimension or numerical scale from dominating the aggregate.

\subsection{Linear Probing Comparison}
\label{sec:app_linear_probe}

We use linear probing to verify that the selected representations expose class-level semantics. Features are precomputed from the frozen encoder for the 1,281,167 training and 50,000 validation images of ImageNet-1K. A single linear layer with 1,000 outputs is then trained using SGD with momentum $0.9$, zero weight decay, a cosine learning-rate schedule, and 300 epochs. We sweep learning rates $\{0.01,0.1\}$ and report the checkpoint with the highest validation top-1 accuracy. For context, we also probe pooled LPIPS-VGG features and a flattened Stable-Diffusion-2.1 VAE latent.

\begin{table}[t]
\centering
\caption{\textbf{ImageNet-1K linear probing with frozen representations.} The five VFMs used in our VLM-based evaluation appear above the separator.}
\label{tab:app_linear_probe}
\small
\setlength{\tabcolsep}{5pt}
\begin{tabular}{lccc}
\toprule
Representation & Feature & Top-1 (\%) & Top-5 (\%) \\
\midrule
ConvNeXt-v2 & spatial average & 86.47 & 97.98 \\
DINOv2 & class token & 85.99 & 97.44 \\
SigLIP2 & attention pool & 85.56 & 97.78 \\
CLIP & class token & 83.27 & 97.14 \\
Inception-v3 & spatial average & 77.73 & 93.89 \\
\midrule
LPIPS-VGG~\citep{zhang2018lpips} & pooled multilevel features & 66.09 & 86.73 \\
SD2.1 VAE~\citep{rombach2022ldm} & flattened spatial latent & 5.62 & 13.74 \\
\bottomrule
\end{tabular}
\end{table}

All five evaluation encoders provide substantially more linearly accessible category information than the reconstruction-oriented references. DINOv2, SigLIP2, and CLIP obtain high accuracy through self-supervised or vision--language training, while ConvNeXt-v2 and Inception-v3 contribute features shaped by supervised recognition. The result confirms that each selected space contains strong high-level information and motivates averaging across models with different inductive biases. The very low linear separability of the VAE latent is also consistent with the DiffC analysis in Sec.~\ref{sec:semantic_collapse_analysis}.

\section{Evaluation Details with Qwen3.5-9B}
\label{sec:app_vlm}

\subsection{Detailed Evaluation Protocol}
\label{sec:app_vlm_protocol}

The VLM evaluation deliberately separates the source-blind questions ``what is recognizable?'' and ``how well formed is it?'' from the source-aware question ``does it preserve the input meaning?''. Table~\ref{tab:app_vlm_stages} summarizes the three calls. The reference inventory is cached once per source image and reused for every codec and bitrate.

\begin{table*}[t]
\centering
\caption{\textbf{Three-stage VLM evaluation.} A semantic unit may describe a scene, entity, identity-defining attribute, action or state, relation or layout, meaningful text or symbol, or context. Degradation itself is recorded as quality evidence rather than as a semantic unit.}
\label{tab:app_vlm_stages}
\small
\setlength{\tabcolsep}{3pt}
\begin{tabular}{p{0.05\textwidth} p{0.20\textwidth} p{0.30\textwidth} p{0.34\textwidth}}
\toprule
Stage & Input to the VLM & Structured output & Role \\
\midrule
A & Source image only & Global interpretation and at most 12 nonredundant units, each with importance $w_r\in\{1,\ldots,5\}$ & Defines source semantics independently for all metrics. \\
B & Reconstruction only & Visible-content description and at most 12 units with importance, identity confidence, and four intrinsic-quality ratings & Produces source-blind evidence for SR and SQ without reference. \\
C & Text inventories from A and B only & Global similarity and one match for every source and reconstruction semantic unit & Measures source coverage in both directions; images and the SR/SQ confidence and quality are withheld. \\
\bottomrule
\end{tabular}
\end{table*}

For each reconstruction unit $c$, identity confidence $p_c\in[0,100]$ measures whether an unprimed observer can defend its stated identity. Four integer ratings in $[0,10]$ assess semantic clarity, structural integrity, appearance naturalness, and artifact non-dominance. Their weighted intrinsic quality is
\begin{equation}
 q_c=10\cdot\left(0.3\cdot q_c^{\mathrm{clarity}}+0.3\cdot q_c^{\mathrm{structure}}
 +0.2\cdot q_c^{\mathrm{naturalness}}+0.2\cdot q_c^{\mathrm{artifact}}\right).
 \label{eq:app_vlm_quality}
\end{equation}
For the reconstruction inventory $\mathcal C_i$, raw semantic recognizability and quality are
\begin{equation}
 \mathrm{SR}^{\mathrm{raw}}_i=
 \frac{\sum_{c\in\mathcal C_i}w_cp_c}{\sum_{c\in\mathcal C_i}w_c},
 \qquad
 \mathrm{SQ}^{\mathrm{raw}}_i=
 \frac{\sum_{c\in\mathcal C_i}w_cq_c}{\sum_{c\in\mathcal C_i}w_c},
 \label{eq:app_vlm_sr_sq}
\end{equation}
with both scores set to zero when no semantic unit is recognizable. SR therefore reflects identity confidence, whereas SQ measures the integrity of the recognized content rather than fidelity to the source or aesthetic preference.

For source units $\mathcal R_i$, the text-only matcher supplies a similarity for each source-to-reconstruction and reconstruction-to-source match. The two directions yield importance-weighted semantic recall and precision,
\begin{equation}
 \rho_i=\frac{\sum_{r\in\mathcal R_i}w_r s(r,\mathcal C_i)}
                  {\sum_{r\in\mathcal R_i}w_r},
 \qquad
 \pi_i=\frac{\sum_{c\in\mathcal C_i}w_c s(c,\mathcal R_i)}
                 {\sum_{c\in\mathcal C_i}w_c}.
 \label{eq:app_vlm_recall_precision}
\end{equation}
Recall penalizes missing source content, while precision penalizes invented or substituted reconstruction content. Their harmonic mean is combined with global scene similarity $G_i$:
\begin{equation}
 F_i=\frac{2\rho_i\pi_i}{\rho_i+\pi_i},
 \qquad
 \mathrm{SC}_i=0.4\cdot G_i+0.6\cdot F_i.
 \label{eq:app_vlm_sc}
\end{equation}
Empty inventories use fixed deterministic rules. In particular, a nonempty source paired with an empty reconstruction inventory receives zero unit-level consistency.

\textbf{JPEG-identity normalization.} The VLM does not use the nominal $[0,100]$ range uniformly across image content: even a clean source may score below 100, while severe degradation may receive nonzero confidence. We therefore evaluate two image-matched controls with the same reconstruction-only prompt. For $\mathrm{M}\in\{\mathrm{SR},\mathrm{SQ}\}$, the identity anchor $\mathrm{U}_i^\mathrm{M}$ is obtained by submitting the clean source as an anonymous candidate; the lower anchor $\mathrm{L}_i^\mathrm{M}$ is obtained from the same source compressed with baseline JPEG at quality 1 and 4:2:0 chroma subsampling:
\begin{equation}
 \mathrm{SR}_i=100\cdot\operatorname{\textbf{clip}}\!\left[
 \frac{\mathrm{SR}_i^{\mathrm{raw}}-\mathrm{L}_i^\mathrm{SR}}{\mathrm{U}_i^\mathrm{SR}-\mathrm{L}_i^\mathrm{SR}},0,1\right],
 \qquad
 \mathrm{SQ}_i=100\cdot\operatorname{\textbf{clip}}\!\left[
 \frac{\mathrm{SQ}_i^{\mathrm{raw}}-\mathrm{L}_i^\mathrm{SQ}}{\mathrm{U}_i^\mathrm{SQ}-\mathrm{L}_i^\mathrm{SQ}},0,1\right].
 \label{eq:app_ji10}
\end{equation}

The frozen evaluator uses Qwen3.5-9B with greedy decoding and schema-constrained JSON output. Codec identity, bitrate, filename, conventional metrics, and other methods' outputs are never supplied to the model. We retain the raw scores, normalized scores, SC subcomponents, and the fractions clipped at either anchor. The three metrics are intentionally independent: a natural but source-unrelated image can obtain high SR and SQ but low SC. These model-based judgments are used as a large-scale diagnostic rather than a replacement for human evaluation.

\begin{figure*}[t]
    \centering
    \includegraphics[width=\textwidth]{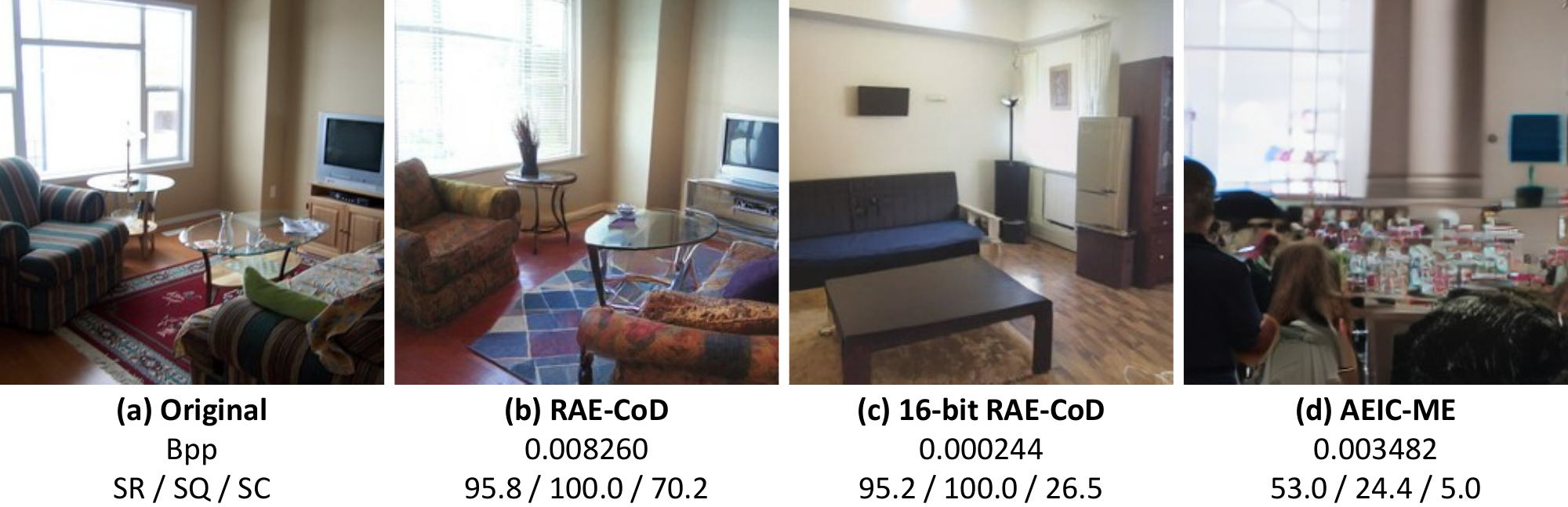}
    \caption{\textbf{Qualitative examples at $256\times256$ for the VLM evaluation.} RAE-CoD at 0.008260 bpp preserves the room layout and principal objects. Its 16-bit reconstruction remains coherent and recognizable but replaces much of the source-specific furniture. AEIC-ME at 0.003482 bpp instead produces a severely degraded, source-unrelated scene, yielding low SR, SQ, and SC.}
    \label{fig:app_vlm_examples}
\end{figure*}

\begin{table*}[t]
\centering
\caption{\textbf{Candidate-unit outputs for the reconstructions in Fig.~\ref{fig:app_vlm_examples}.} The quality tuple is ordered as (clarity, structure, naturalness, artifact non-dominance).}
\label{tab:app_vlm_example_units}
\footnotesize
\setlength{\tabcolsep}{1.0pt}
\renewcommand{\arraystretch}{0.90}
\begin{tabular}{c c p{0.68\textwidth} c c c}
\toprule
Case & Unit & Reconstruction-only semantic description & $w_c$ & $p_c$ & Quality tuple $\rightarrow q_c$ \\
\midrule
\multirow{10}{*}{(b)}
 & C1 & A living room interior with furniture arranged around a central coffee table & 5 & 95 & $(6,7,6,4)\rightarrow59$ \\
 & C2 & A patterned armchair with a floral or paisley design in earth tones & 4 & 90 & $(6,7,6,5)\rightarrow61$ \\
 & C3 & A glass-topped coffee table with a metal base in the center of the room & 4 & 90 & $(6,7,6,5)\rightarrow61$ \\
 & C4 & A large area rug with a geometric pattern of blue, red, and white squares & 3 & 85 & $(6,7,6,5)\rightarrow61$ \\
 & C5 & A small, round, dark-colored side table with curved legs near the window & 3 & 85 & $(6,7,6,5)\rightarrow61$ \\
 & C6 & A television set on a low cabinet in the background right & 3 & 80 & $(5,6,5,5)\rightarrow53$ \\
 & C7 & A large window with horizontal blinds on the left wall & 2 & 90 & $(7,8,8,7)\rightarrow75$ \\
 & C8 & A vase containing dried branches or twigs on the side table & 2 & 75 & $(5,6,6,5)\rightarrow55$ \\
 & C9 & A partial view of a matching patterned sofa in the right foreground & 2 & 80 & $(6,7,6,5)\rightarrow61$ \\
 & C10 & Reddish-brown flooring, likely wood or laminate & 1 & 70 & $(5,6,5,5)\rightarrow53$ \\
\midrule
\multirow{10}{*}{(c)}
 & C1 & A sparsely furnished living room or studio apartment interior & 5 & 95 & $(6,7,7,4)\rightarrow61$ \\
 & C2 & A dark, rectangular coffee table in the foreground & 4 & 90 & $(7,8,8,7)\rightarrow75$ \\
 & C3 & A dark futon or sofa bed with a blue mattress cover & 4 & 90 & $(7,8,8,7)\rightarrow75$ \\
 & C4 & A tall, beige refrigerator standing on a dark pedestal & 3 & 85 & $(6,7,7,6)\rightarrow65$ \\
 & C5 & A tall, dark wooden cabinet or hutch & 3 & 80 & $(6,7,7,6)\rightarrow65$ \\
 & C6 & A black floor lamp in the corner & 2 & 80 & $(6,7,7,6)\rightarrow65$ \\
 & C7 & A window with sheer white curtains & 2 & 85 & $(6,7,7,6)\rightarrow65$ \\
 & C8 & A small, dark rectangular object mounted on the wall & 2 & 60 & $(5,6,6,6)\rightarrow57$ \\
 & C9 & Light-colored walls and wood-laminate flooring & 2 & 90 & $(7,8,7,7)\rightarrow73$ \\
 & C10 & Flash-photography lighting & 1 & 90 & $(8,8,5,7)\rightarrow72$ \\
\midrule
\multirow{6}{*}{(d)}
 & C1 & A crowded indoor public space, likely a retail or exhibition area & 5 & 60 & $(3,4,4,2)\rightarrow33$ \\
 & C2 & A group of people seen from behind and looking toward the display & 4 & 70 & $(3,4,4,3)\rightarrow35$ \\
 & C3 & A long reflective counter containing many small, indistinct colorful items & 4 & 60 & $(2,4,3,3)\rightarrow30$ \\
 & C4 & A large, light-colored structural pillar in the middle ground & 3 & 80 & $(4,6,5,6)\rightarrow52$ \\
 & C5 & A large background window admitting bright, overexposed light & 2 & 80 & $(3,5,4,5)\rightarrow42$ \\
 & C6 & A blue square object mounted on the background wall & 2 & 70 & $(3,5,4,5)\rightarrow42$ \\
\bottomrule
\end{tabular}
\end{table*}

\subsection{Qualitative Examples}
\label{sec:app_vlm_examples}

We use image \texttt{MSCOCO-30K/000000002347.png} to contrast three reconstruction outcomes: a source-preserving RAE-CoD reconstruction, a recognizable but source-divergent RAE-CoD reconstruction at 16 bits, and a semantically collapsed AEIC-ME reconstruction with low recognizability, quality, and consistency. Fig.~\ref{fig:app_vlm_examples} reports the final SR, SQ and SC scores. The unit-level evidence underlying the raw scores is summarized in Table~\ref{tab:app_vlm_example_units}.

\textbf{JPEG-identity normalization for this source.} For SR, the identity and JPEG anchors are $(\mathrm{U}^{\mathrm{SR}},\mathrm{L}^{\mathrm{SR}})=(88.148,45.294)$; for SQ, $(\mathrm{U}^{\mathrm{SQ}},\mathrm{L}^{\mathrm{SQ}})=(53.778,32.176)$. The $(\mathrm{SR}^{\mathrm{raw}},\mathrm{SQ}^{\mathrm{raw}})$ pairs for RAE-CoD, 16-bit RAE-CoD, and AEIC-ME are $(86.379,60.103)$, $(86.071,67.393)$, and $(68.000,37.450)$, respectively. Per-image normalization maps them to the final pairs $(95.872,100.000)$, $(95.153,100.000)$, and $(52.985,24.414)$ shown in Fig.~\ref{fig:app_vlm_examples}. The two RAE-CoD SQ values exceed the clean-image SQ anchor and are therefore clipped to 100, whereas the AEIC-ME value lies between the JPEG and identity anchors.

\textbf{RAE-CoD: source-preserving reconstruction.} The source inventory describes a naturally lit living room containing striped seating, a glass coffee table, a red patterned rug, a television and stand, and a side table with a lamp. At 0.008260 bpp, RAE-CoD retains the room type, principal furniture, and overall layout, although some colors and smaller attributes change. After JPEG-identity normalization, its final scores are $\mathrm{SR}=95.872$ and $\mathrm{SQ}=100.000$. The source-to-candidate similarities are $(85,60,95,55,90,45,90,85,0,0)$, and the reverse similarities are $(85,60,95,55,90,90,90,40,60,85)$. These outputs give global similarity $G=65.000$, recall $\rho=71.111$, precision $\pi=76.379$, and $F=73.651$, producing $\mathrm{SC}=70.191$.

\textbf{16-bit RAE-CoD: high-quality but source-divergent reconstruction.}
At 16 bits ($0.000244$~bpp), the reconstruction-only inventory still recognizes a coherent living room or studio apartment, including a coffee table, futon, refrigerator, cabinet, lamp, and window. It therefore retains high final source-blind scores, with $\mathrm{SR}=95.153$ and $\mathrm{SQ}=100.000$. However, most source-specific content has been replaced: the source-to-candidate similarities fall to $(40,30,30,0,20,40,50,60,60,0)$, while the reverse similarities are $(40,30,30,0,20,40,50,20,60,0)$. Consequently, $G=20.000$, $\rho=31.852$, $\pi=30.000$, and $F=30.898$, giving $\mathrm{SC}=26.539$.

\textbf{AEIC-ME: low-quality and source-inconsistent reconstruction.} Although AEIC-ME uses 0.003482 bpp, more than fourteen times the rate of the 16-bit RAE-CoD, its output is interpreted as a crowded retail or exhibition space containing several people, a display counter, a pillar, and a window. Heavy blur, block, lost detail, and color smearing make the semantic units difficult to identify and poorly formed. Its final scores fall to $\mathrm{SR}=52.985$ and $\mathrm{SQ}=24.414$. The only unit-level overlap is the generic presence of a window: the source-to-candidate similarities are $(0,0,0,0,0,0,20,0,0,0)$ and the reverse similarities are $(0,0,0,0,20,0)$. Together with $G=10.000$, this gives $\rho=1.481$, $\pi=2.000$, $F=1.702$, and $\mathrm{SC}=5.021$. The example therefore captures semantic collapse: the reconstruction is both difficult to recognize as coherent content and largely unrelated to the source, despite receiving substantially more bits than the 16-bit RAE-CoD.

\section{Reverse-Channel Coding with DiffC}
\label{sec:app_diffc}

\subsection{Algorithm}
\label{sec:app_diffc_algorithm}

DiffC turns the variational structure of a pretrained diffusion model into a progressive lossy code~\citep{theis2022lossy,vonderfecht2025diffc}. Let $x_0$ denote the datum in the model's native space and consider the variance-preserving forward marginal
\begin{equation}
 q(x_t\mid x_0)=\mathcal N\!\left(\sqrt{\bar\alpha_t}x_0,
 \bar\beta_t I\right),\qquad \bar\beta_t=1-\bar\alpha_t.
 \label{eq:app_diffc_forward}
\end{equation}
For a transition from noise level $t$ to a cleaner level $s<t$, let $\alpha_{t\mid s}=\bar\alpha_t/\bar\alpha_s$ and $\beta_{t\mid s}=1-\alpha_{t\mid s}$. The tractable Gaussian bridge is
\begin{equation}
 q(x_s\mid x_t,x_0)=\mathcal N(A_{s,t}x_0+B_{s,t}x_t,\sigma_{s,t}^{2}I),
 \label{eq:app_diffc_posterior}
\end{equation}
where
\begin{equation}
 A_{s,t}=\frac{\sqrt{\bar\alpha_s}\,\beta_{t\mid s}}{\bar\beta_t},\quad
 B_{s,t}=\frac{\sqrt{\alpha_{t\mid s}}\,\bar\beta_s}{\bar\beta_t},\quad
 \sigma_{s,t}^{2}=\frac{\bar\beta_s}{\bar\beta_t}\beta_{t\mid s}.
 \label{eq:app_diffc_coefficients}
\end{equation}
The encoder knows $x_0$, whereas both encoder and decoder can evaluate the diffusion prediction $\hat{x}_0=f_\theta(x_t,t)$ from their shared state. Replacing $x_0$ by $\hat{x}_0$ defines the model proposal
\begin{equation}
 p_\theta(x_s\mid x_t)=\mathcal N(A_{s,t}\hat{x}_0+B_{s,t}x_t,\sigma_{s,t}^{2}I).
 \label{eq:app_diffc_proposal}
\end{equation}
After standardizing by the common covariance, the proposal and target are $P=\mathcal N(0,I)$ and $Q=\mathcal N(m,I)$, where
\begin{equation}
 m=\frac{A_{s,t}}{\sigma_{s,t}}(x_0-\hat{x}_0),\qquad
 D_{\mathrm{KL}}(Q\Vert P)=\frac{\lVert m\rVert_2^2}{2\ln2}\ \text{bits}.
 \label{eq:app_diffc_kl}
\end{equation}
Thus, a better diffusion prediction directly reduces the information needed to communicate the next posterior state.

Reverse-channel coding communicates a draw from $Q$ when both parties know $P$ and share pseudorandomness. In the Poisson functional representation (PFR) used by DiffC, both parties generate candidates $z_n\sim P$ and exponential arrival times $T_n$. The encoder selects
\begin{equation}
 n^*=\arg\min_n T_n\frac{p(z_n)}{q(z_n)}
 \label{eq:app_pfr}
\end{equation}
and transmits only the winning index; the decoder regenerates the same sequence and recovers $z_{n^*}$. Its expected description length is close to $D_{\mathrm{KL}}(Q\Vert P)$, with logarithmic overhead. Since candidate search grows exponentially with the number of bits, the implementation partitions the latent coordinates into independent chunks, limits each chunk to a small bit budget, and arithmetic-codes the winner indices under the induced Zipf model.

Starting from shared $x_T\sim\mathcal N(0,I)$, the encoder progressively communicates posterior samples toward $x_0$. In the idealized scheme, the cumulative rate down to a stopping time $t^*$ is
\begin{equation}
 R(t^*)\simeq D_{\mathrm{KL}}\!\left(q(x_T\mid x_0)\Vert p(x_T)\right)
 +\sum_{(t,s):T\rightarrow t^*}
 D_{\mathrm{KL}}\!\left(q(x_s\mid x_t,x_0)\Vert p_\theta(x_s\mid x_t)\right).
 \label{eq:app_diffc_rate}
\end{equation}
Exact coding continues to $x_0$. For lossy compression, DiffC stops at $x_{t^*}$ and completes the trajectory with deterministic probability-flow denoising. Stopping earlier transmits fewer bits and delegates more decisions to the generative prior; the code is progressive because each additional transition refines the previously shared state.

The three models evaluated in Sec.~\ref{sec:semantic_collapse_analysis} are trained with a linear flow path $y_t=(1-t)x_0+t\epsilon$. We convert it to the variance-preserving (VP) form in Eq.~\eqref{eq:app_diffc_forward} by
\begin{equation}
 n(t)=\sqrt{(1-t)^2+t^2},\qquad
 x_t=\frac{y_t}{n(t)},\qquad
 \bar\alpha_t=\frac{(1-t)^2}{n(t)^2},\quad
 \bar\beta_t=\frac{t^2}{n(t)^2}.
 \label{eq:app_flow_to_vp}
\end{equation}
The denoiser is evaluated in its native flow coordinates to obtain $\hat{x}_0$; the common Gaussian bridge and reverse-channel coder can then be applied without retraining.

\subsection{DiffC on Different Diffusion Models}
\label{sec:app_diffc_models}

We wrap three pretrained class-conditioned generators with the same DiffC pipeline. The evaluation set is a balanced subset of 5,000 ImageNet validation images, containing five center-cropped $256\times256$ images from each of the 1,000 classes. The source class is available to both parties and is charged as a fixed 10-bit message, or $10/256^2=0.000153$ bpp. For each image, progressive encoding is run once to the maximum rate, and the state nearest each target rate is retained. The reported coding rate consists of the arithmetic-coded PFR payload and the class label. File headers and stored experimental timestep/KL metadata are excluded consistently for all three models.

For Pixel-DiT, the source itself, scaled to $[-1,1]$, is the diffusion datum. For VAE-DiT, we use a deterministically seeded sample from the frozen VAE posterior; for RAE-DiT, we use the normalized DINOv3 representation. Since all three generators use rectified flow, Eq.~\eqref{eq:app_flow_to_vp} supplies a common VP process for reverse-channel coding. After the selected noisy state is recovered, each model follows its native Euler trajectory and released guidance setting. These controls isolate the effect of diffusion space while preserving the model-specific sampler required for valid generation.

\begin{figure*}[t]
    \centering
    \includegraphics[width=\textwidth]{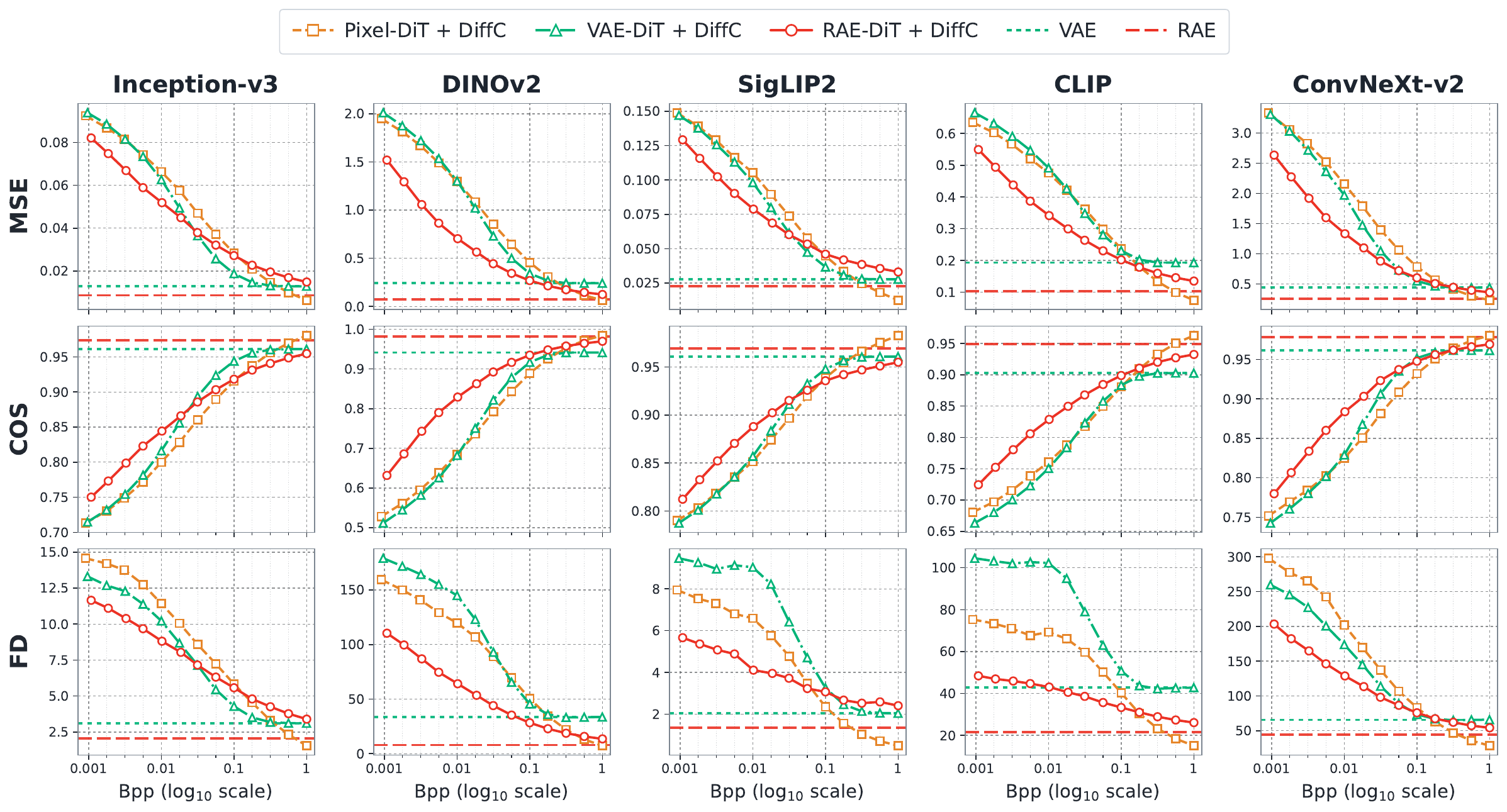}
    \caption{\textbf{Unaggregated semantic results for the DiffC space comparison.} Rows report feature MSE, cosine similarity, and Fr\'echet Distance; columns correspond to the five VFMs used by our aggregate metrics. Horizontal lines denote the reconstruction ceilings of the frozen VAE and RAE. These are the per-representation results underlying Fig.~\ref{fig:diffusion_space_quantitative}.}
    \label{fig:app_diffc_semantic}
\end{figure*}

Fig.~\ref{fig:app_diffc_semantic} gives the corresponding unaggregated semantic values for Fig.~\ref{fig:diffusion_space_visualization}. Across all five VFMs,  RAE-DiT has the best feature MSE, cosine similarity and Fr\'echet Distance throughout the extreme-rate range up to approximately $0.056$ bpp. At higher rates, Pixel-DiT eventually overtakes it because direct pixel generation has no autoencoder reconstruction ceiling.

\subsection{One-Step DiffC for 16-bit RAE-CoD}
\label{sec:app_one_step_diffc}

The 16-bit RAE-CoD condition constrains broad content but does not determine a unique reconstruction, so different initial noise samples can change identity or layout. The one-step experiment in Sec.~\ref{sec:discussion_ablation} adds a single target-dependent transition near the pure-noise endpoint while leaving RAE-CoD unchanged. The encoder forms the clean normalized RAEv2 target $x_0$ and the decoded codec condition $c$. Both sides start from the same seeded $x_1\sim\mathcal N(0,I)$ and evaluate the conditioned denoiser to obtain $\hat{x}_0=f_\theta(x_1,1,c)$. For a destination time $s<1$, the posterior and proposal simplify to
\begin{equation}
 q(x_s\mid x_0)=\mathcal N(\sqrt{\bar\alpha_s}x_0,\bar\beta_sI),
 \qquad
 p(x_s\mid x_1,c)=\mathcal N(\sqrt{\bar\alpha_s}\hat{x}_0,\bar\beta_sI),
 \label{eq:app_one_step_distributions}
\end{equation}
so the ideal additional rate is
\begin{equation}
 R_{\mathrm{DiffC}}=
 \frac{1}{2\ln2}\left(\frac{1-s}{s}\right)^2
 \lVert x_0-\hat{x}_0\rVert_2^2 \quad\text{bits}.
 \label{eq:app_one_step_rate}
\end{equation}
A chunked PFR search selects a source-compatible noisy state, the receiver reproduces it from shared randomness and the winner indices, and the original 100-step shifted Euler sampler of RAE-CoD resumes at $s$. The matched no-DiffC baseline uses the same codec condition, initial noise, and sampler but starts directly at $t=1$.

The approximately four extra bits quoted in Sec.~\ref{sec:discussion_ablation} are the ideal KL quantity in Eq.~\eqref{eq:app_one_step_rate}. The current diagnostic executes both PFR sides and verifies exact equality of the recovered noisy states, but passes the winner indices and KL metadata in memory. It does not yet serialize a standalone DiffC fragment or count its arithmetic-coder termination, metadata, byte-alignment, and container overhead. The 20-bit point should therefore be interpreted as a theoretical-rate analysis showing that very little additional source information can reduce sampling ambiguity, rather than as the measured size of a deployable file.

\section{Additional Implementation Details}
\label{sec:app_implementation}

\subsection{Model Details}
\label{sec:app_model_details}

\begin{table*}[t]
\centering
\caption{\textbf{RAE-CoD architecture at $256\times256$.} Spatial sizes omit the batch dimension.}
\label{tab:app_model_shapes}
\footnotesize
\setlength{\tabcolsep}{3pt}
\renewcommand{\arraystretch}{1.2}
\begin{tabular}{l p{0.6\textwidth} l}
\toprule
Component & Construction & Output \\
\midrule
RAEv2 encoder & DINOv3-L/16 with seven-layer aggregation and normalization & $x_0:1024\times16\times16$ \\
Pixel encoder & Four $2\times$ downsampling stages with channels $64,128,192,256$ & $256\times16\times16$ \\
Latent encoder & Concatenate both maps, project, and downsample once & $y:256\times8\times8$ \\
Hyper encoder & Two $2\times$ downsampling stages & $z:256\times2\times2$ \\
VQ bottleneck & 4-bit (16-entry) vector codebook & $\hat z:256\times2\times2$ \\
Hyper decoder & Two $2\times$ upsampling stages to produce a hyperprior & $256\times2\times2$ \\
Entropy model & Scalar quantization with 4-step quadtree-partition spatial context and a Gaussian entropy model & $\hat y:256\times8\times8$ \\
Latent decoder & Fuse $\hat z$ and $\hat y$, and two $2\times$ upsampling stages & $c:1152\times16\times16$ \\
Auxiliary head & Token-wise MLP $1152\rightarrow1024\rightarrow1024$ & 256 tokens \\
RAEv2 DDT & 28 transformer encoder blocks (width 1440, 20 heads) and 2 transformer decoder blocks (width 2048, 16 heads) & $\hat{x}_0:1024\times16\times16$ \\
RAEv2 decoder & 28-layer transformer, width 1152, 16 heads, patch size 16 & $3\times256\times256$ \\
\bottomrule
\end{tabular}
\end{table*}

Table~\ref{tab:app_model_shapes} traces RAE-CoD at $256\times256$ resolution. The frozen RAEv2 encoder adopts the DINOv3-L/16 encoder and extracts layers $\{11,13,15,17,19,21,23\}$, averages their patch tokens, adds a broadcast global mean from the last selected layer, and applies the pretrained RAEv2 channel-wise normalization. This produces $x_0\in\mathbb R^{1024\times16\times16}$. The pixel encoder uses Inception-style depthwise convolutions and gated channel mixing~\citep{zhang2025stablecodec}. Its output is concatenated with the spatial DINOv3 map before producing $y$. The four spatial masks decode complementary subsets of $y$ sequentially; at each pass, the hyperprior and previously decoded coefficients predict a Gaussian mean and scale for the next subset. The reconstructed main latent $\hat y$ and hyperprior are concatenated and decoded into 256 condition tokens. The final codec contains approximately 57.2M parameters. Its auxiliary head is used only for training and aligns conditions with the DINOv3 targets. The RAEv2 DDT contains approximately 875.4M parameters, and the RAEv2 decoder remains frozen.

\subsection{Training Details}
\label{sec:app_training_details}

Training uses 23.2M images from ImageNet-21K and CC12M, resized and center-cropped to $256\times256$. The RAEv2 encoder and decoder remain frozen, while the codec is learned from scratch. Diffusion time is sampled by drawing $u=\operatorname{sigmoid}(\xi)$ for $\xi\sim\mathcal N(0,1)$ and applying $t=8u/[1+7u]$. The conditioning input is dropped with probability 0.1 to train the unconditional branch used by guidance. The full DDT output and the early output after encoder block 8 are trained against the same flow target. We use $\lambda_{\mathrm{REPA}}=1$, $\lambda_{\mathrm{VQ}}=0.25$, and $\lambda_{\mathrm{aux}}=0.5$.

\begin{table*}[t]
\centering
\caption{\textbf{RAE-CoD training and inference configuration.}}
\label{tab:app_training_inference}
\footnotesize
\setlength{\tabcolsep}{3pt}
\renewcommand{\arraystretch}{1.2}
\begin{tabular}{l p{0.37\textwidth} p{0.37\textwidth}}
\toprule
Setting & Stage I & Stage II / inference \\
\midrule
Initialization & Pretrained RAEv2 encoder, DDT and decoder; new codec, DDT LoRAs and condition projection & Corresponding Stage-I rate checkpoint \\
Trainable parameters & Codec, LoRA ($r=32$), condition projection, and output interfaces & Codec and all DDT parameters \\
Frozen modules & RAEv2 encoder, decoder and the base DDT parameters & RAEv2 encoder and decoder \\
Rate weight $\lambda_{\mathrm{rate}}$ & Undergoes 0.1, 2, 12, 16, 24, 32, and 48 & Fixed in $\{12,16,24,32,48\}$ \\
Rate schedule & 0.1 at step 0, 2 at 20K, then 12, 16, 24, 32, 48 at 30K intervals starting at 30K & 100K steps for each operating point \\
Learning rate & $10^{-4}$ & $10^{-5}$ \\
Optimizer & AdamW, zero weight decay & AdamW, zero weight decay \\
Precision & BF16 mixed precision & BF16 mixed precision \\
Effective batch & 128 & 128 \\
GPU hardware & Four NVIDIA A100 GPUs & Four NVIDIA A100 GPUs \\
EMA & Decay 0.9995 & Decay 0.9995 \\
Sampling & -- & 100-step Euler; time shift 8; internal guidance 1.78 on $t\in[0.1,1]$; no CFG \\
\bottomrule
\end{tabular}
\end{table*}

During Stage~I, LoRA is inserted into the attention, feed-forward, adaptive-normalization, and output projections of the pretrained DDT. The base parameters remain frozen while the new codec, condition projection, and LoRA parameters adapt to compression. After the initial warm-up, the increasing rate weight progressively removes information from $y$. During Stage~II, the appropriate Stage-I checkpoint initializes each target rate, the LoRA weights are merged, and the complete DDT is optimized jointly with the codec. This separation avoids abruptly perturbing the pretrained prior when compression training begins.

\subsection{Inference Details}
\label{sec:app_inference_details}

We evaluate the EMA model weights. The hyper latent $z$ is replaced by its nearest entry in the 4-bit codebook (16-entry), and the main latent $y$ is rounded using means predicted from the hyperprior and previously decoded partitions. The reported rate for a $256\times256$ image is
\begin{equation}
 R=\frac{-\log_2 p(\hat y)+16}{256^2}\quad\text{bpp},
 \label{eq:app_raecod_rate}
\end{equation}
where the 16-bit term is the exact fixed-length hyper latent payload. The decoded condition is flattened into 256 tokens and reused at every diffusion step. We initialize a $1024\times16\times16$ Gaussian state, integrate the shifted Euler trajectory for 100 steps, and apply internal guidance as $x_{\mathrm{base}}+1.78\cdot(x_{\mathrm{full}}-x_{\mathrm{base}})$ over $t\in[0.1,1]$ following~\citep{singh2026raev2}. Classifier-free guidance is disabled. Finally, the frozen RAE decoder maps the generated representation back to RGB pixels.

\section{Additional Comparison on Kodak}
\label{sec:app_kodak}
\begin{figure*}[t]
    \centering
    \includegraphics[width=\textwidth]{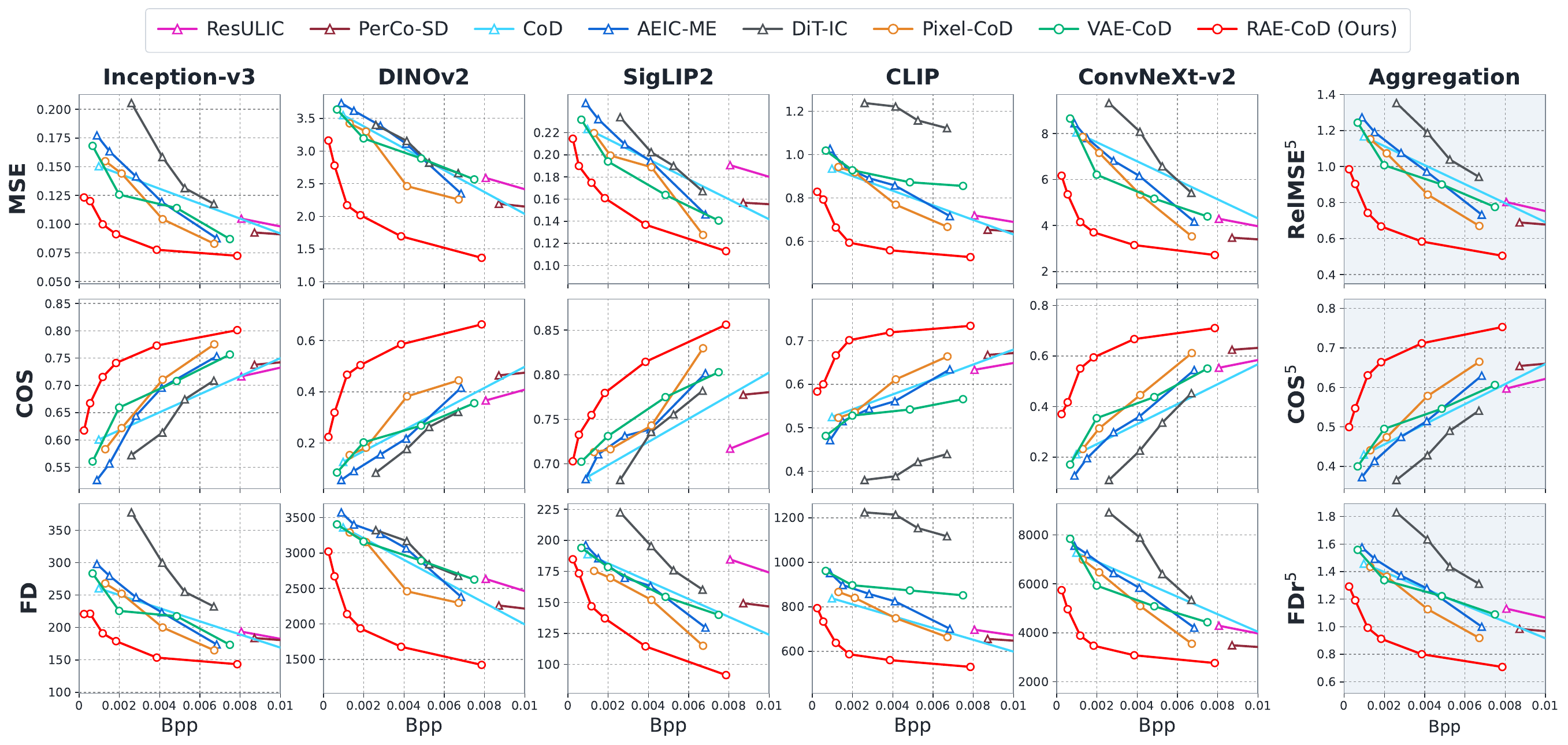}
    \caption{\textbf{VFM-based semantic evaluation on Kodak at $256\times256$.} We report feature MSE, cosine similarity, and Fr\'echet Distance in five VFM spaces, together with their aggregate metrics.}
    \label{fig:app_rd_kodak}
\end{figure*}

Fig.~\ref{fig:app_rd_kodak} extends the VFM comparison to Kodak. RAE-CoD traces the strongest overall rate-semantic envelope, obtaining lower feature MSE and Fr\'echet Distance and higher cosine similarity than the alternatives at comparable rates across the aggregate metrics and individual VFM spaces. This advantage persists toward the 16-bit endpoint.

\section{Additional Visualization}
\label{sec:app_additional_visualization}
\begin{figure*}[t]
    \centering
    \includegraphics[width=\textwidth]{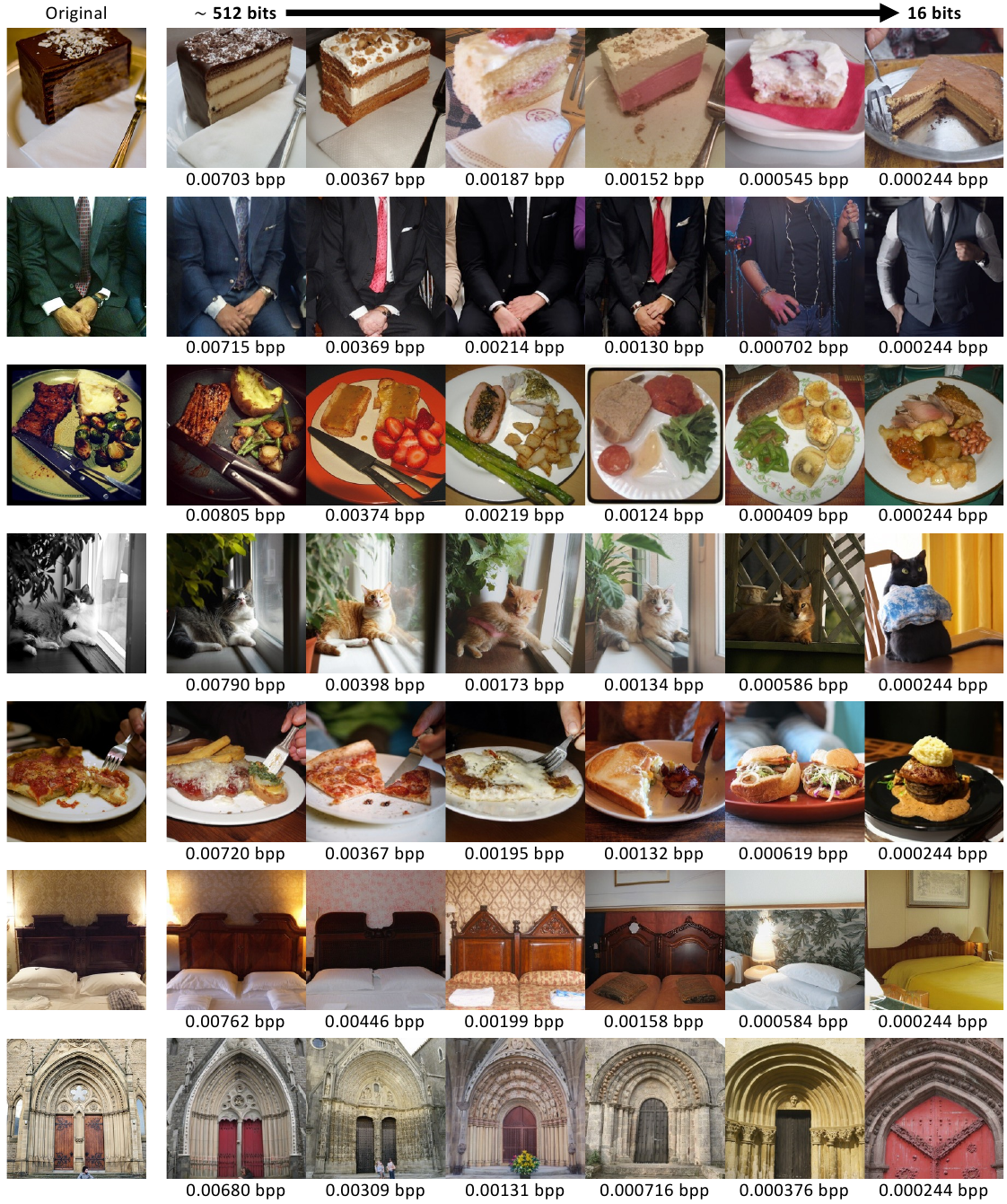}
    \caption{\textbf{Additional visual examples of RAE-CoD on MSCOCO-30K at $256\times256$.} From approximately 512 bits to 16 bits (0.000244 bpp), the reconstructions exhibit graceful semantic erosion from the source while remaining visually coherent.}
    \label{fig:large_fig}
\end{figure*}

Fig.~\ref{fig:large_fig} shows the progressive visual behavior of RAE-CoD on more examples. As the rate decreases from roughly $0.007$--$0.008$ bpp to 16 bits, fine appearance and source-specific attributes are lost first, followed by changes in object identity and spatial arrangement. Nevertheless, the outputs remain recognizable and naturally structured even when their correspondence to the source becomes weak. The repeated transition across examples visually supports the intended shift from faithful reconstruction toward unconditional generation, rather than an abrupt collapse into malformed content.

\end{document}